\pdfoutput=1

\documentclass[11pt]{article}

\usepackage[final]{acl}

\usepackage{natbib}
\usepackage{xcolor}
\usepackage{hyperref}
\usepackage{times}
\usepackage{latexsym}
\usepackage{multirow}
\usepackage{multicol}
\usepackage{graphicx}
\usepackage{siunitx}
\usepackage{wrapfig}
\usepackage[most]{tcolorbox}
\usepackage[T1]{fontenc}

\usepackage[utf8]{inputenc}

\usepackage{microtype}

\usepackage{inconsolata}

\usepackage{booktabs} 
\usepackage{amsmath}
\usepackage{mathtools}
\usepackage{multirow}
\usepackage{siunitx}
\usepackage{lineno}
\usepackage{enumitem}
\usepackage{amsfonts}       
\usepackage{nicefrac}       
\usepackage{amsmath}
\usepackage{amssymb}
\usepackage[noend]{algpseudocode}
\usepackage{algorithm}
\usepackage{widebar}
\usepackage{subcaption}
\usepackage[textsize=scriptsize]{todonotes}
\usepackage{cuted}

\usepackage{array} 

\definecolor{todoblue}{RGB}{33,150,243} 

\definecolor{pink-color}{RGB}{255,1,97} 

\definecolor{red-color}{RGB}{255,111,105} 
\definecolor{dark-red-color}{RGB}{174,0,1} 
\definecolor{light-red-color}{RGB}{255,152,150}

\definecolor{green-color}{RGB}{136,216,176} 
\definecolor{dark-green-color}{RGB}{49,163,84} 
\definecolor{light-green-color}{RGB}{136,216,176} 
\definecolor{sea-green-color}{RGB}{95,158,160} 

\definecolor{dark-blue-color}{RGB}{49,130,189} 
\definecolor{light-blue-color}{RGB}{158,202,225} 

\definecolor{grey-color}{RGB}{167,173,186}
\definecolor{dark-grey-color}{RGB}{79,91,102}
\definecolor{gold-color}{RGB}{247,205,15} 

\newcommand{\gold}{\mbox{\textsc{Gold}}}
\newcommand{\fewgen}{\mbox{\textsc{FewGen}}}

\newcommand{\ToI}{\mbox{\textsc{ToI}}}

\newcommand{\DistilBERT}{\mbox{\textsc{DistilBERT}}}
\newcommand{\BERT}{\mbox{\textsc{BERT}}}

\newcommand{\GPTTwoXL}{\mbox{\textsc{GPT2-XL}}}
\newcommand{\ChatGPT}{\mbox{\textsc{GPT3.5-Turbo}}}
\newcommand{\ChatGPTShort}{\mbox{\textsc{GPT3.5-T}}}
\newcommand{\PhiMini}{\mbox{\textsc{Phi-3 Mini}}}
\newcommand{\Mixtral}{\mbox{\textsc{Mixtral}}}

\newcommand{\aref}[1]{\hyperref[#1]{Appendix~\ref*{#1}}}

\newcommand{\ZeroGen}{\textsc{ZeroGen}}

\newcommand{\SunGen}{\textsc{SunGen}}
\newcommand{\ReGen}{\textsc{ReGen}}
\newcommand{\SynthesizRR}{\textsc{SynthesizRR}}
\newcommand{\LetsSynth}{\textsc{S3}}
\newcommand{\AttrPrompt}{\textsc{AttrPrompt}}

\newcommand{\synthd}{\ensuremath{\mathbin{
    \mathcal{D}_{\textsc{Synth}}
}}}

\newcommand{\higherbetter}{\ensuremath{(\uparrow)}}
\newcommand{\lowerbetter}{\ensuremath{(\downarrow)}}

\newcommand{\insection}[2][.]{{\setlength{\parskip}{6pt} \noindent\textbf{#2#1}}}

\newcommand{\ignore}[1]{}
\newcommand{\ig}[1]{}

\newcommand{\promptsubsection}[1]{
\setlength{\parskip}{6pt} \noindent\textbf{{#1}:}
}
\newcommand{\param}[1]{
\textcolor{pink-color}{\scriptsize{\texttt{\detokenize{#1}}}}
}

\newcommand{\prompttext}[1]{``\textcolor{dark-grey-color}{#1}''}

\newtcolorbox[list inside=prompt]{prompt}[1][]{
    colbacktitle=black!60,
    coltitle=white,
    fontupper=\footnotesize,
    boxsep=5pt,
    left=0pt,
    right=0pt,
    top=0pt,
    bottom=0pt,
    boxrule=1pt,
    title={Prompt~\thetcbcounter}, 
    #1,
}

\newcommand{\urlsmall}[1]{{\scriptsize{\url{\detokenize{#1}}}}}

\newcolumntype{h}{>{\setbox0=\hbox\bgroup}c<{\egroup}@{}}
\newcolumntype{C}[1]{>{\centering\arraybackslash}m{#1}}
\newcolumntype{R}[1]{>{\raggedleft\arraybackslash}p{#1}}
\newcolumntype{L}[1]{>{\raggedright\arraybackslash}p{#1}}

\newcommand{\cellhalign}[1]{\multicolumn{1}{c}{#1}}

\newcommand{\LLaMa}{\mbox{\textsc{LLaMa2}}}

\newcommand{\IMDb}{\mbox{\textsc{IMDb}}}

\newcommand{\AGNews}{\mbox{\textsc{AG News}}}
\newcommand{\AG}{\mbox{\textsc{AG.}}}

\newcommand{\ToIHeadlines}{\mbox{\textsc{ToI Headlines}}}

\newcommand{\Humor}{\mbox{\textsc{Humor}}}

\newcommand{\Hum}{\mbox{\textsc{Hum.}}}

\newcommand{\distas}[1]{\mathbin{\overset{#1}{\kern0pt\sim}}}%
\def\corrsyn{{\textsc{CorrSynth}}}

\def\corrsynreallyshort{{\textsc{Corr}}}

\def\cX{\mathcal{X}}
\def\cY{\mathcal{Y}}

\def\cV{\mathcal{V}}
\def\cS{\mathcal{S}}
\def\cI{\mathcal{I}}
\def\cT{\mathcal{T}}
\def\eos{\mathtt{<eos>}}
\def\Prompt{\mathtt{prompt}}
\def\lg{\mathtt{lg}}
\def\mP{P}
\newcommand{\mbf}[1]{\mathbf{#1}}

\newcommand{\Ex}[1]{\mathbb{E}\left[#1 \right]}

\newcommand\sk[1]{
}
\newcommand\vm[1]{
}
\newcommand\ad[1]{
}

\title{CorrSynth - A Correlated Sampling Method for Diverse Dataset Generation from LLMs}

\makeatletter
\renewcommand{\@fnsymbol}[1]{\ifcase#1\or *\else\@arabic{#1}\fi}
\makeatother

\author{
\textbf{Suhas S Kowshik}\thanks{Equal contribution: order was determined by random dice rolls. Correspondence to: \texttt{adivekar@amazon.com}},
\textbf{Abhishek Divekar}\textsuperscript{*}, 
\textbf{Vijit Malik}
\\
Amazon
\\
\texttt{\{kowssuhp, adivekar, vijitvm\}@amazon.com}
}

\begin{document}
\maketitle

\begin{abstract}

Large language models (LLMs) have demonstrated remarkable performance in diverse tasks using zero-shot and few-shot prompting. Even though their capabilities of data synthesis have been studied well in recent years, the generated data suffers from a lack of diversity, less adherence to the prompt, and potential biases that creep into the data from the generator model. In this work, we tackle the challenge of generating datasets with high diversity, upon which a student model is trained for downstream tasks. Taking the route of decoding-time guidance-based approaches, we propose \corrsyn{}, which generates data that is more diverse and faithful to the input prompt using a correlated sampling strategy. Further, our method overcomes the complexity drawbacks of some other guidance-based techniques like classifier-based guidance. With extensive experiments, we show the effectiveness of our approach and substantiate our claims. In particular, we perform intrinsic evaluation to show the improvements in diversity. Our experiments show that \corrsyn{} improves both student metrics and intrinsic metrics upon competitive baselines across four datasets, showing the innate advantage of our method.
\end{abstract}

\section{Introduction}
\ad{TODO talk about LLMs here, not BERT immediately} Pretrained language models (LLMs) \citep{devlin-etal-2019-bert} have achieved strong performance on text classification with a large amount of task-specific training data. However, in real world scenarios, collecting labeled data can be challenging due to expense and need for domain expertise.
Recently, several works have focused on generating texts using versatile LLMs such as GPT-4 \citep{Achiam2023GPT4TR}, Claude \citep{Bai2022TrainingAH}, Mistral \citep{jiang2023mistral}, Mixtal~\citep{jiang2024mixtral} and subsequently distill a student model on the synthetically generated data \citep{west-etal-2022-symbolic}. However, generated datasets suffer from a lack of diversity \citep{yu2023large} and regurgitate the biases of the teacher LLMs, which proliferate into the student model. Although prior works have utilized retrieval augmented generation for diverse dataset synthesis \citep{divekar2024synthesizrr}, here we focus on the more fundamental challenge of improving or controlling generations \emph{given a prompt and context}.

\begin{figure}[t!]
\centering
    \includegraphics[width=0.95\linewidth]{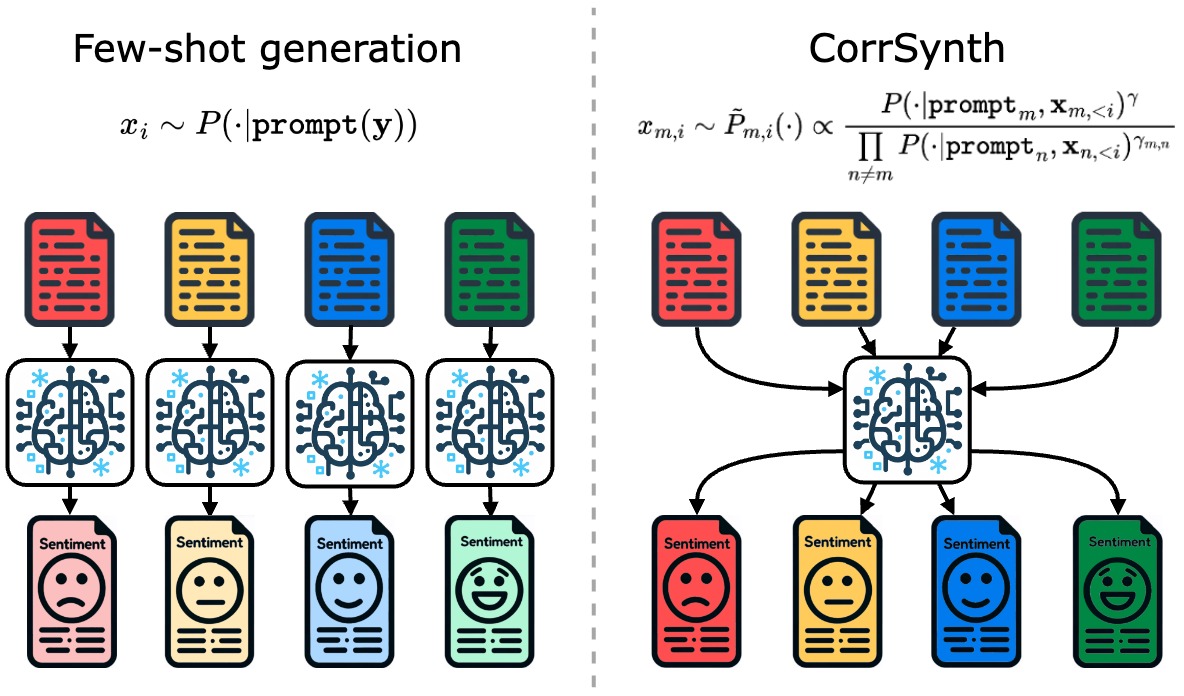}
    \caption{\corrsyn\ introduces anti-correlation \mbox{between} examples, compared to few-shot generation.}
    \label{fig:corrsynth_high_level}
\vspace{-3ex}
\end{figure}

In particular, we focus on synthetic data generation for supervised text classification tasks and take the route of decoding time guidance based approaches \citep{sanchez2023stay,o2023contrastive,li2023contrastive,chuang2023dola}, which aim to tackle the challenge of improving diversity and faithfulness to target class in these generated datasets. Motivated by recent works on Classifier Free Guidance (CFG) \citep{sanchez2023stay}, we introduce a novel guidance based strategy, \corrsyn.\ad{State more clearly why we want to use CFG/guidance based approaches for dataset synthesis, and the problems of CFG (move that up here from later in the para).}
In \corrsyn{}, generations are kept faithful to the synthesis instruction, while introducing greater diversity and similarity to human text. \corrsyn{} is a correlated sampling approach which generates multiple sequences in parallel with strong inter-dependence between them. 
\ad{Too much }The main idea is as follows: when generating an instance of a particular class and sampling the next token, we \emph{contrast} its logits with logits corresponding to partially generated instances from other classes.
This is a simple but crucial change compared to CFG: in \corrsyn, the contrasting logits for a class/label are obtained from generations corresponding to other labels, whereas in CFG, the contrasting logits are obtained feeding back the generation for the current label into the LLM with prompts corresponding to other labels. To synthesize a $K$--class classification dataset, this requires $K$--times fewer forward passes compared to CFG. Furthermore, we can smoothly trade-off diversity and improve class-separability \ad{needs slightly more detail, seems to short}by introducing contrasts between logits from the same or different classes.


In summary, our contributions are: \mbox{(1) we} develop a general correlated sampling approach, \corrsyn{}, that can generate multiple correlated sequences in parallel from an LLM, by explicitly introducing \textit{contrasts} between parallel generations during the sampling of each token, (2) we apply this to classification dataset synthesis, with the goal of improving diversity of synthetic generations, (3) we demonstrate how our method overcomes the limitations of CFG and controllable synthesis in regards to diversity and label-separation, (4) we benchmark our approach on tasks ranging from humor detection, sentiment analysis and topic classification in regional news headlines. Our intrinsic analysis find that \corrsyn{} generates datasets with higher representation of tail entities, lexical diversity and similarity to human text, and distillation accuracy of student models, compared to four state of the art baselines. 

\section{Background}
\label{sec:background}

\paragraph{Notation:} For $n\in\mathbb{N}$, let $[n]=\{1,2,\cdots,n\}$. An LLM is defined through its vocabulary $\cV$ and the auto-regressive sequence distribution $\mP$ or equivalently the logits $\lg$. Let $\cV^*=\cup_{n\geq 1}\cV^n$ denote the space of all finite sequences of tokens from $\cV$. We denote sequences of tokens from $\cV$ using lower boldface letters like $\mathbf{u},\mathbf{v}$. For any sequence of tokens $\mathbf{w}=(w_1,\cdots,w_n)\in \cV^n$ from $\cV$, and any $j\in[n]$, let $\mathbf{w}_{<j}=(w_1,\cdots,w_{j-1})$ if $j>1$, else, it is an empty sequence. Similarly $\mathbf{w}_{\leq j}=(w_1,\cdots,w_j)$. For any two sequences $\mathbf{u},\mathbf{v}\in\cV^*$ let $(\mathbf{u},\mathbf{v})$ denote their concatenation. We denote by $\mP(\mathbf{v}|\mathbf{u})$ the conditional probability of generating $(\mathbf{u},\mathbf{v})$ given that $\mathbf{u}$ has already been generated i.e., probability that $\mathbf{v}$ is a continuation of $\mathbf{u}$ for a given $\mathbf{u}$. Furthermore, for any $\mathbf{u},\mathbf{v}\in \cV^*$, we use $\mP(\cdot|\mathbf{u},\mathbf{v})$ to denote the conditioning on the concatenation $(\mathbf{u},\mathbf{v})$. For any prompt $\Prompt\in\cV^*$ , and any $\mathbf{w}\in\cV^n$, the auto-regressive distribution $\mP$ satisfies
\begin{align*}
    &\mP(\mathbf{w}|\Prompt)=\nonumber\\
    &\mP(w_1|\Prompt)\prod_{j=2}^{n}\mP(w_j|\Prompt,w_1,\cdots,w_{j-1})
\end{align*}
When we describe natural language domains using $\cX$, $\cY$ we mean either in the sense of them containing natural language sentences or as subsets of $\cV^*$, it will be clear from the context.

We consider dataset generation for text classification tasks. Suppose we have a multiclass text classification problem with $K$ classes as $[K]$ and input domain $\cX$. Let $\cY=\{\mbf{y}_1,\cdots,\mbf{y}_K\}$ be the space of label verbalizations for the $K$ classes i.e., $\mbf{y}_k$ is a textual description of label $k\in[K]$. A natural language example input is denoted as $\mbf{x}\in\cX$. So the learning problem is defined on $\cX\times \cY$: given a data generating distribution $P_{XY}$ on $\cX\times \cY$ the task is to learn a classifier $h:\cX\to \cY$ (using some training data) such that $\Ex{l(h(\mbf{x}),\mbf{y})}$ is minimized for a given loss function $l:\cY\times\cY\to \mathbb{R}$, where the expectation is taken with respect to $P_{XY}$. 

Given the rapid advancement of LLMs like GPT-4, Llama2, Mistral etc. we are interested in utilizing the world knowledge and reasoning capabilities of these large models to generate synthetic training data for the textual $K$-class classification problem. Similar to recent works in this domain \cite{ye2022zerogen,gao2022self,Meng2022GeneratingTD,meng2023tuning,yu2023regen,ye2022progen,yu2024large,guo2024generative}, we consider the setup of prompting teacher LLM with a prompt $\Prompt$ that includes a label $\mbf{y}\in\cY$, a few In-Context Learning (ICL) examples for the label $\mbf{y}$ and potentially any other instance dependent attributes, and the prompt tasks the LLM to generate a synthetic instance $\mbf{x}\in \cX$ whose true label is expected to be $\mbf{y}$ i.e., the aim is to generate $x\distas{}P_{X|Y=\mbf{y}}$. That is, we generate a synthetic dataset \synthd{}. A student language model (e.g., a BERT-style pre-trained encoder model \citep{devlin-etal-2019-bert}) is trained on \synthd{}. 

For the ICL examples, we assume that we have access to a \emph{seed set} of examples $\mathcal{D}_{\textsc{Seed}} = \{(\mbf{x}_1,\mbf{y}_1),\ldots,(\mbf{x}_n,\mbf{y}_n)\}$. For us, typically $n$ is such that we have around $50$ examples per class.\sk{Abhishek, is this correct?} We assume that $\mathcal{D}_{\textsc{Seed}}$ is not large enough to train an effective student, but instead a larger synthetic dataset $\mathcal{D}_{\textsc{Synth}}=\{ (\tilde{\mbf{x}}_i,\mbf{y}_i)\}_{i=1}^m$  will be needed.

A standard approach to dataset synthesis is few shot generation i.e. \fewgen{} \cite{NEURIPS2020_1457c0d6,ye2022progen,Yehudai2024GenieAH}. For instance, consider a task of detecting a business news article. In order to synthesize a dataset for this task, we could prompt the LLM appropriately, include few ICL examples. The LLM might generate a fairly decent article. But when we sample a large number of generations we see that the there is lack of diversity: similar entities are repeated, popular topics are highlighted and potential stylistic differences from a human written text. These could affect the performance of a student model that is trained on such dataset.

A ``good'' synthetic dataset must ensure that the conditional distribution of instances given any label must closely approximate that of the true distribution $P_{XY}$. This includes: i) correct semantic separation of labels, ii) preservation of intra-label semantic diversity and of course, iii) fluent and coherent generations. In order to achieve (i) and (ii) (without compromising on (iii)), we present a method, \corrsyn{}, in the flavor of decoding time guidance techniques \cite{li2023contrastive,o2023contrastive,sanchez2023stay,chuang2023dola}. In these works, at inference time, the token probability distribution is tilted by another distribution obtained either from a different LLM, or same LLM with a different prompt, or different layers of the same LLM. In particular, we take inspiration from the classifier free guidance~\cite{ho2021classifierfree} method applied to text based LLMs \cite{sanchez2023stay}. 
\corrsyn{} aims to control i) diversity in generations, ii) similarity to human crafted gold dataset, iii) cross label separation and at the same time iv) improve the student performance. The core idea of our approach is to perform correlated or dependent sampling from the LLM i.e., multiple sequences are generated in parallel that have strong dependency between each other. \autoref{fig:corrsynth_high_level} illustrates our method. More details are given in \autoref{sec:method}. This method can be used in conjunction with other synthetic dataset generation approaches like retrieval augmented generation~\cite{lewis2020retrieval}.

\section{Method}
\label{sec:method}
Now we describe our novel \corrsyn{} method of sampling from an LLM. Although it is a general technique, we choose to motivate it from the perspective of data synthesis for a text based supervised learning problem. 



\subsection{\corrsyn}
\label{sec:corrsyn-intro}
Let us consider the case of binary classification with verbalized labels $\{\mathbf{y}_0,\mathbf{y}_1\}$. As is standard in dataset synthesis~\cite{ye2022zerogen,gpt3}, we create class-conditioned prompt $\Prompt(\mathbf{y})$ which describes the task using verbalization $\mathbf{y}\in\{\mathbf{y}_0,\mathbf{y}_1\}$, and prompt the LLM to generate continuations as our synthetic input $x$. In-context examples are used to guide the generations to follow the format specified in the prompt. 
Suppose we want to generate two instances $\mbf{x},\bar{\mbf{x}}$ corresponding to labels $\mbf{y}, \bar{\mbf{y}}$ respectively. In \corrsyn we generate them together as follows. Let $0\leq \delta\leq \gamma$. Then:
\begin{align}
    & x_i \distas{}  \tilde \mP_i(\cdot) \propto \frac{P(\cdot|\Prompt(\mathbf{y}),\mathbf{x}_{<i})^\gamma}{\mP(\cdot|\Prompt(\mathbf{\bar{\mathbf{y}}}),\bar{\mathbf{x}}_{<i})^{\gamma-\delta}} \label{eq:2corr_eq1}\\
   & \bar{x}_i  \distas{} \tilde Q_i(\cdot) \propto \frac{\mP(\cdot|\Prompt(\mathbf{\bar{\mathbf{y}}}),\bar{\mathbf{x}}_{<i})^\gamma}{\mP(\cdot|\Prompt(\mathbf{y}),\mathbf{x}_{<i})^{\gamma-\delta}} \label{eq:2corr_eq2}
\end{align}
We hypothesize that the sequences $\mbf{x}, \bar{\mbf{x}}$ generated auto-regressively using equations \eqref{eq:2corr_eq1} and \eqref{eq:2corr_eq2} are naturally anti-correlated: they tend to be far apart in the embedding space of the LLM. This is because, when sampling a token  for a sequence, the plausible tokens for the contrasting sequences are weighted down. Furthermore, at token $i$, even if the numerator and denominator distributions in \eqref{eq:2corr_eq1} highlight different entities or parts of speech, we expect the overall semantic meaning to be weakly present in the individual token distributions due to the attention mechanism. Thus even at these tokens, we posit that the contrast provides a signal that moves the generated sequences apart. This reasoning is based on intuition that requires careful experiments to prove. Nonetheless, we will demonstrate this separation of sequences in our analysis in \autoref{sec:analysis}. So we call the sequences $\mbf{x},\bar{\mbf{x}}$ to be contrastive to one another. We can use this property to control label separation as well as intra-class diversity when generating synthetic instances.

\insection[]{Crucial change from CFG}: in denominator of \eqref{eq:2corr_eq1}, the conditioned partial sequence $\bar{\mbf{x}}_{<i}$ is actually expected to be faithful to $\Prompt(\bar{\mbf{y}})$, and thus the effect of guidance would persist even after many tokens. Additionally, we generate two sequences together, leading to a two fold increase in the number of forward passes compared to a single generation, whereas CFG would require four times more. We introduce another parameter $\delta$ which controls the strength of the denominator contrast. More details on CFG for dataset synthesis are in \autoref{sec:CFG}. 


\subsection{$M$--\corrsyn}
\label{sec:M-corrsyn}
Next, we generalize from binary to $M$-way contrastive generation. Suppose we have $M$ prompts $\{\Prompt_1,\cdots,\Prompt_M\}$. We want to generate $M$ sequences $\{\mbf{x}_m:m\in [M]\}$ such that $\mbf{x}_m$ is faithful to $\Prompt_m$. Let $\gamma>0$ be the guidance, and let $0\leq \delta\leq \gamma$. We introduce $M^2$ weights $\{\gamma_{m,n}:m,n\in[M], \gamma_{m,m}=0\}$. We generate the $i$-th token of $\mbf{x}_m=(x_{m,1},\cdots,x_{m,n_m}), \forall m$:
\begin{align}
     x_{m,i}&\distas{}\tilde \mP_{m,i}(\cdot)\nonumber\\
    &\propto \frac{\mP(\cdot| \Prompt_m,\mbf{x}_{m,<i})^{\gamma}}{\prod_{n \neq m} \mP(\cdot| \Prompt_n,\mbf{x}_{n,<i})^{\gamma_{m,n}}}\label{eq:M-Corr-1}
\end{align}
Next we describe our choice of $\gamma_{m,n}$. 

\subsubsection{Uniform contrastive guidance}
\label{sec:unif_guidance}
We set a parameter $\delta$ that controls the total amount of contrast guidance: for each $m$, $\sum_n \gamma_{m,n}=\gamma-\delta$. Then, when generating $i$-th token for $\mbf{x}_m$, we set $\gamma_{m,n}=0$ for sequences $\mbf{x}_{n}$ that have already hit the EOS token. Then, we uniformly divide $\gamma-\delta$ among remaining $\gamma_{m,n}$\footnote{$\gamma_{m,n}$ also depends on the token index $i$; we suppress it.}. More details are in \aref{sec:app_unif_guidance}. Using uniform contrastive guidance, $M$-\corrsyn{} has a natural geometric mean interpretation that we discuss in \aref{sec:geometric}.

\subsubsection{\corrsyn{} for $K$-class synthesis}
Now we briefly describe how we use \corrsyn{} in data synthesis for $K$ class classification. Recall that in $K$-class classification problem over $\cX\times\cY$ we have classes $[K]$ with label verbalizations $\{\mbf{y}_1,\cdots,\mbf{y}_K\}$. To generates instances for each class, we create prompts as follows. Let $R\in\mathbb{N}$ be the repeat factor. In $M$-\corrsyn{}, we take $M=KR$, and prompts in $\{\Prompt_{m}:m=(k-1)R+r,\, 1\leq r\leq R\}$ correspond to class $k$ for all $k\in[K]$. For $m=(k-1)R+r$, prompt $\Prompt_{m}$ asks the LLM to generate instances for class $k$ contains positive ICL examples for that class. These ICL examples differ across $r$. Thus in equation~\eqref{eq:M-Corr-1}, a generation for class $k$ is, potentially, contrasted against the remaining $R-1$ generations from the same class, as well as the $(K-1)R$ generations from other classes. Based on setting the weights $\gamma_{m,n}$ to be zero for either intra-label terms or cross label terms, we get three scenarios:
\insection[]{\corrsyn{} Cross-label}: When generating a sequence for class $k$ and $m=(k-1)R+r$, we set $\gamma_{m,n}=0$ for $n\in\{(k-1)R+r':r'\neq r\}$. So only terms belonging to classes $k'\neq k$ appear in the denominator of \eqref{eq:M-Corr-1}.\\
\insection[]{\corrsyn{} Intra-label}:  When generating a sequence for class $k$ and $m=(k-1)R+r$, we set $\gamma_{m,n}=0$ for $n\in\{(k'-1)R+r':r'\in[R],k'\neq k\}$. So only terms belonging to class $k$ appear in the denominator of \eqref{eq:M-Corr-1}.\\
\insection[]{\corrsyn{} Hybrid}: denominator of \eqref{eq:M-Corr-1} contains terms that belong to the same class as well as those that belong to other classes. We separately set the target guidance for each of the Cross- and Intra-label terms: we fix two targets $\gamma_{intra}$ and $\gamma_{cross}$ such the sum of $\gamma_{m,n}$ for Intra and Cross label terms are set to $\gamma_{intra}$ and $\gamma_{cross}$ respectively. Then we uniformly split the target guidances $\gamma_{intra}$ and $\gamma_{cross}$ in respective groups. More details of $K$-class \corrsyn{} is given in \aref{sec:K-corrsyn}

\subsubsection{Logits Space computation}
The \corrsyn{} method is implemented using vector arithmetic in the space of LLM outputs i.e. logits space. Complete details are in \aref{sec:logit_corrsyn}. Taking logarithm of the \corrsyn{} sampling equations gives us similar results\footnote{Caveat: taking logarithm gives us log-probabilities which are normalized version of logits. Experimentally, we have not found significant impact of this normalization.}.

\subsubsection{Plausibility Constraint ($\alpha$)}

The contrast terms in \corrsyn{} could sometimes upweight irrelevant tokens i.e. those which are not plausible conditioned on the prompt/label under consideration. To mitigate this, we borrow the idea of plausibility constraint from \cite{li2023contrastive, o2023contrastive} to limit the token up weighting space: by reducing the space of logits to those tokens that have at least $\alpha$ fraction of the mode of the numerator distribution in \eqref{eq:M-Corr-1}. We provide the complete formulation in \aref{sec:plaus}.
 
\section{Experimental Setup}
\label{sec:expt-setup}

\begin{table}
\small
\centering
\begin{tabular}{@{}lccc@{}}
\toprule
\textbf{Dataset} & \textbf{Type} & \textbf{Class} & \textbf{Train, Test} \\ 
\midrule
\AGNews   & Topic            & $4$              & $115\text{K}, 7.6\text{K}$           \\
\ToIHeadlines   & Topic            & $10$             & $52\text{K}, 10\text{K}$             \\
\Humor   & Sentiment      & $2$     & $15\text{K}, 3\text{K}$   \\
\IMDb   & Sentiment      & $2$     & $20\text{K}, 25\text{K}$   \\ 
\bottomrule
\end{tabular}
\caption{Dataset statistics.}
\vspace{-1em}
\label{tab:tasks}
\end{table}




\paragraph{Datasets.} \ad{Remove and cut down} We experiment on 4 datasets described in \autoref{tab:tasks}, which are selected to encompass a wide scope of generation tasks (news headlines, news articles, humorous product questions and movie reviews). Previous work primarily benchmarked only sentiment and topic classification datasets. We consider: (1) \AGNews{}~\citep{zhang2015character}, a popular topic classification dataset where each news summary is mapped to a news topic. The generation task involves generating news summaries based on news topics; (2) \ToIHeadlines{}\cite{toiheadlines}, similarly is a topic classification dataset of regional news headlines in India that maps news topics to news headlines; the generation task is similar to \AGNews{}. The difficulty is that the headlines is regionalized to Indian news and hence requires India specific entities; (3) \Humor~\citep{humor} task involves generating humorous and non-humorous questions from retrieved product details; (4) \IMDb{} \cite{maas-etal-2011-learning} is a sentiment task with binary labels. Prompts are in \autoref{app:prompts}.

\begin{table*}[h]
\centering
\begin{tabular}{
L{65pt}
c 
C{16pt} 
C{17pt} 
C{18pt} 
C{24pt} 
C{20pt} 
|C{16pt} 
C{17pt} 
C{18pt} 
C{24pt} 
C{20pt} 
}

\toprule
\multirow{2}{*}{\textbf{Method}}   
& \multirow{2}{*}{\textbf{Teacher}} 
& \multicolumn{4}{c}{\textbf{Accuracy \higherbetter}}
& \multirow{2}{*}{\textbf{Avg.}}
& \multicolumn{4}{c}{\textbf{MAUVE \higherbetter}}
& \multirow{2}{*}{\textbf{Avg.}}
\\ 
\cmidrule(l){3-6}         
\cmidrule(l){8-11}         
& \textbf{LM} 
& \textbf{\AG} 
& \textbf{\ToI} 
& \textbf{\Hum} 
& \textbf{\IMDb} 
&
& \textbf{\AG} 
& \textbf{\ToI} 
& \textbf{\Hum} 
& \textbf{\IMDb} 
&
\\ 
\midrule
\gold          
& \multicolumn{1}{c}{-}                
& 91.4         & 78.9         & 92.9          & 91.4 & 88.7
& -         & -         & -          & - & -
\\ 
\midrule
\multicolumn{12}{c}{\underline{\textsc{In-Context Learning}}} \\
[0.5ex]
\fewgen 
& \multicolumn{1}{c}{\PhiMini}          
& 83.8         & 69.7         & 68.5          & 85.1 & 76.8
& 91.0         & 86.3        & 83.7          & 67.7 & 82.2
\\ 
\fewgen 
& \multicolumn{1}{c}{\Mixtral}          
& 72.3         & 47.3         & 82.8          & 87.1 & 67.5
& 87.1         & 91.6         & 87.0          & 64.6 & 82.6
\\ 
[1.0ex]

\corrsynreallyshort-Intra 
& \multicolumn{1}{c}{\PhiMini}          
& 84.8         & 71.0         & 84.7          & 87.1 & 81.9
& 82.3         & 83.2         & 82.3          & 77.4 & 81.3
\\ 
\corrsynreallyshort-Hybrid 
& \multicolumn{1}{c}{\PhiMini}          
& \textbf{85.1}         & \textbf{71.1}         & 85.1          & 86.8 & \textbf{82.1}
& 77.5         & 82.0         & 81.7          & 71.0 & 78.1
\\ 
[0.5ex]

\corrsynreallyshort-Intra 
& \multicolumn{1}{c}{\Mixtral}
& 78.5         & 68.9         & \textbf{86.5}          & \textbf{88.6} & 80.1
& \textbf{94.4}         & 95.6         & 95.5          & 76.8 & 90.1
\\  
\corrsynreallyshort-Hybrid 
& \multicolumn{1}{c}{\Mixtral}
& 73.6         & 68.4         & 86.0          & 88.1 & 79.0
& 93.8         & \textbf{96.1}         & \textbf{97.1}          & \textbf{80.5} & \textbf{91.9}
\\ 
[1.0ex]
\toprule
\multirow{2}{*}{\textbf{Method}}   
& \multirow{2}{*}{\textbf{Teacher}} 
& \multicolumn{4}{c}{\textbf{Self-BLEU-5 \lowerbetter}}
& \multirow{2}{*}{\textbf{Avg.}}
& \multicolumn{4}{c}{\textbf{Entity-Entropy \higherbetter}}     
& \multirow{2}{*}{\textbf{Avg.}}
\\
\cmidrule(l){3-6}         
\cmidrule(l){8-11}         
& \textbf{LM} 
& \textbf{\AG} 
& \textbf{\ToI} 
& \textbf{\Hum} 
& \textbf{\IMDb} 
&
& \textbf{\AG} 
& \textbf{\ToI} 
& \textbf{\Hum} 
& \textbf{\IMDb} 
&
\\
\midrule
\gold          
& \multicolumn{1}{c}{-}                
& 17.1         & 7.9         & 19.8          & 27.9 & 18.2
& 6.6         & 6.1         & 5.1          & 7.5 & 6.3
\\ 
\midrule
\multicolumn{12}{c}{\underline{\textsc{In-Context Learning}}} \\
[0.5ex]
\fewgen 
& \multicolumn{1}{c}{\PhiMini}          
& 33.9         & 15.3         & 39.9          & 57.7 & 36.7
& 6.6         & 6.3         & 4.3          & 5.3 & 5.6
\\ 
\fewgen 
& \multicolumn{1}{c}{\Mixtral}          
& 39.4         & 37.9         & 64.6          & 66.5 & 52.1
& 5.9         & 5.2         & 3.6         & 5.2 & 5.0
\\ 
[1.0ex]
\corrsynreallyshort-Intra 
& \multicolumn{1}{c}{\PhiMini}          
& 13.1         & 9.0         & 23.5          & 24.9 & 17.6
& \textbf{7.4}         & \textbf{6.9}         & \textbf{4.9}          & \textbf{6.5} & \textbf{6.4}
\\ 
\corrsynreallyshort-Hybrid 
& \multicolumn{1}{c}{\PhiMini}          
& \textbf{12.1}         & \textbf{8.7}         & \textbf{22.8}          & \textbf{19.2} & \textbf{15.7}
& \textbf{7.4}         & \textbf{6.9}        & 4.8          & 6.4 & \textbf{6.4}
\\ 
[0.5ex]
\corrsynreallyshort-Intra 
& \multicolumn{1}{c}{\Mixtral}          
& 18.9         & 17.6         & 45.3          & 33.0 & 28.7
& 6.3          & 5.7         & 3.7          & 6.0 & 5.4
\\  
\corrsynreallyshort-Hybrid 
& \multicolumn{1}{c}{\Mixtral}          
& 17.5         & 18.4         & 41.4          & 27.4 & 26.2
& 6.5         & 5.6         & 4.1         & 6.4 & 5.7
\\ 
\bottomrule
\end{tabular}
\caption{
Evaluation of intrinsic dataset quality and \DistilBERT\ student model fine-tuned on real and synthetic datasets. We report mean accuracy numbers across 5 runs. When generating each instance, we select 3 in-context examples at random to prime the LLM's next-token distribution before sampling continuations. 
}
\vspace{-1em}
\label{tab:accuracy-diversity-icl}
\end{table*}

\begin{figure}[!t]
\centering
\includegraphics[width=0.5\textwidth]{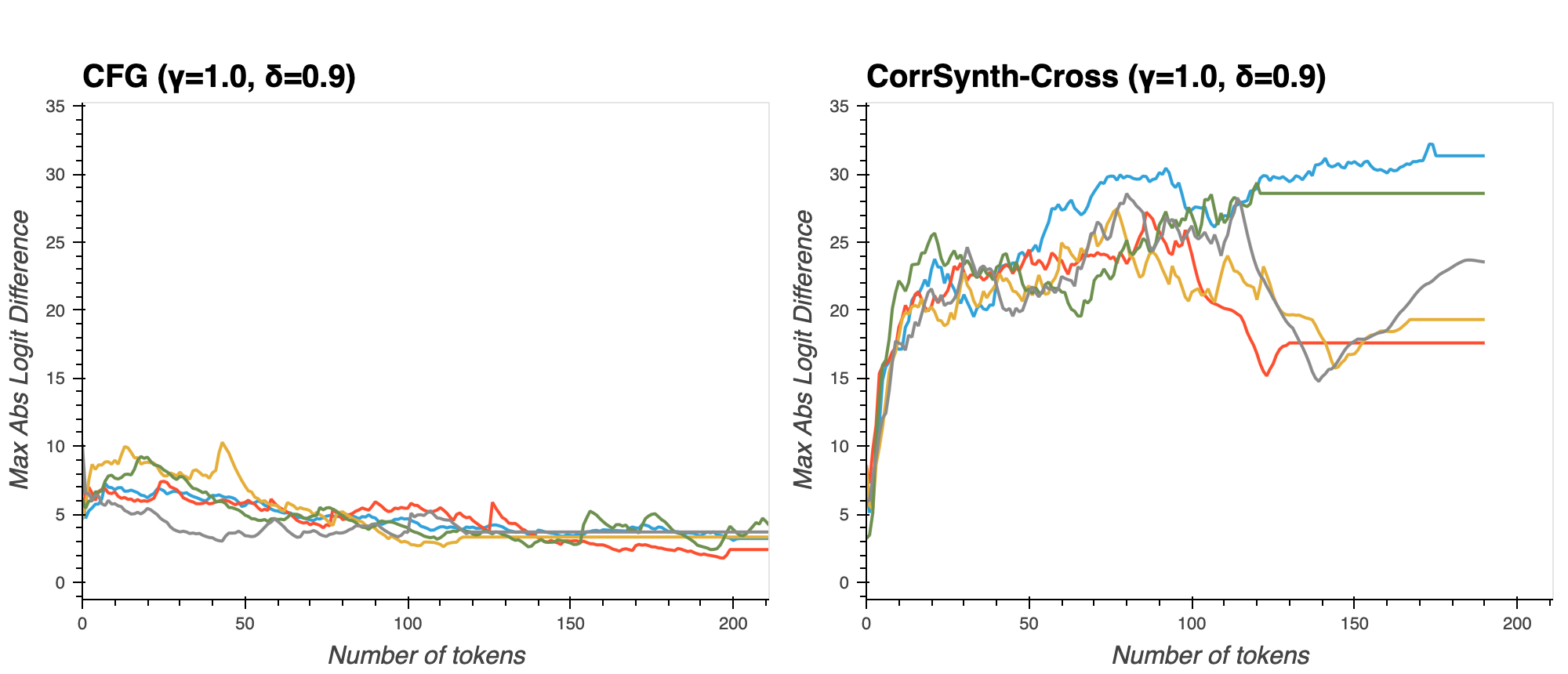}
\vspace{-0.8cm}
\caption{Generation progression from CFG and \corrsyn. We sample five generations using 3-shot prompts from \IMDb. The colored lines represent the absolute difference between logits of the current generation and contrast for each generation timestep (taken as an exponential moving average).}
\vspace{-1em}
\label{fig:cfg_vs_corrsynth}
\end{figure}



\paragraph{Teachers and students.} \ad{TODO fix this to be Mixtral. Cite.} As a teacher model, we use a frozen \Mixtral\ (8x7B) \citep{jiang2024mixtral} or \PhiMini\ (3.8B) \citep{abdin2024phi} for the data generation step. Following \citep{divekar2024synthesizrr}, we select examples randomly from the train set: 50 ICL examples per class for multi-class and 100 per class for binary. We think that this is a reasonable number of labeled examples since we are trading off the effort of labeling versus developing a potential zeroshot technique (which may not work well in practice). We use \DistilBERT{} student model (66M params \citet{Sanh2019DistilBERT}) as it is popular in prior work.

\begin{table*}[h]
\centering
\setlength{\tabcolsep}{3pt}
\begin{tabular}{
L{95pt}
c 
c 
c 
c 
c 
c 
c 
c 
c 
}

\toprule
\multirow{2}{*}{\textbf{Method}}   
& \multirow{2}{*}{\textbf{Teacher}} 
& \multicolumn{2}{c}{\textbf{Accuracy}}
& \multicolumn{2}{c}{\textbf{MAUVE}}
& \multicolumn{2}{c}{\textbf{Self-BLEU-5}}
& \multicolumn{2}{c}{\textbf{Entity-Entropy}}
\\ 
\cmidrule(l){3-4}
\cmidrule(l){5-6}
\cmidrule(l){7-8}
\cmidrule(l){9-10} 
& \textbf{LM} 
& \textbf{{ } \AG} 
& \textbf{\IMDb { }} 
& \textbf{{ } \AG} 
& \textbf{\IMDb { }} 
& \textbf{{ } \AG} 
& \textbf{\IMDb { }} 
& \textbf{{ } \AG} 
& \textbf{\IMDb { }} 
\\ 
\midrule
\gold          
& -
& 91.4 & 91.4
& -	 & -	
& 17.1 & 27.9
& 6.6 & 7.5
\\ 
\midrule
\multicolumn{10}{c}{\underline{\textsc{Retrieval-based methods}}}  \\
\ReGen  
& \BERT
& 82.7 & $\otimes$
& 68.1 & $\otimes$
& 56.5 & $\otimes$
& \textbf{8.1 }& $\otimes$
\\ 
\SynthesizRR
& \LLaMa
& 84.6 & 84.8	
& \textbf{92.6} & 72.6	
& 34.2 & 62.9	
& 7.2 & 5.7
\\
\midrule 
\multicolumn{10}{c}{\underline{\textsc{Non-retrieval methods}}}  \\
\SunGen
& \GPTTwoXL
& $\otimes$ & 84.9	
& $\otimes$ & 68.7	
& $\otimes$ & \textbf{15.4}	
& $\otimes$ & 4.9
\\ 
\LetsSynth
& \ChatGPTShort
& $\otimes$ & \textbf{87.1}	
& $\otimes$ & 62.0	
& $\otimes$ & 62.2	
& $\otimes$ & 5.7
\\ 
\AttrPrompt
& \ChatGPTShort
& 79.8	& $\otimes$	
& 52.8	& $\otimes$	
& 39.8	& $\otimes$	
& 6.0	& $\otimes$
\vspace{1.5ex} 
\\ 
(Ours) \corrsynreallyshort-Intra
& \textsc{Phi-3 Mini}
& 84.8 & \textbf{87.1}
& 82.3 & \textbf{77.4}	
& 13.1 & 24.9	
& 7.4 & \textbf{6.5}
\\ 
(Ours) \corrsynreallyshort-Hybrid 
& \textsc{Phi-3 Mini}
& \textbf{85.1} & 86.8	
& 77.5 & 71.0	
& \textbf{12.1} & 19.2	
& 7.4 & 6.4
\\
\bottomrule
\end{tabular}
\caption{
Comparison of quality metrics and \DistilBERT\ student model fine-tuned on 6k rows from each approach. Mean accuracy across 5 training runs is considered. $\otimes$ indicates datasets were not released by authors.
}
\vspace{-1em}
\label{tab:baselines}
\end{table*}

\paragraph{Evaluation criteria} The task of evaluation of quality of text generation is quite challenging~\citep{llm-eval-survey}. Following prior works like \cite{divekar2024synthesizrr}, we evaluate synthetic generations based on several metrics. \textbf{Self-BLEU} \citep{bleu,zhu2018texygen} measures lexical diversity of a corpus of texts based on $n$-gram overlap between pairs of examples. \textbf{Entity entropy} measures the \textit{diversity} of entities in the generated texts using the distribution of each of 16 entity-types (inferred from a pre-trained named entity recognition model). Dataset which have high occurrence of few popular entities score lower on entropy. \textbf{MAUVE} \citep{liu-etal:divergence:neurips2021} measures closeness to human-written text using representations from a pre-trained \GPTTwoXL{} model. We also measure the \textbf{student accuracy} when trained on the synthetic data. We do not consider label preservation accuracy as it is susceptible to easy examples \cite{divekar2024synthesizrr}. In order to analyse the behavior of our strategy, we also study the label-wise cosine similarity of the generations, low dimensional embeddings of the generations using UMAP~\cite{mcinnes2020umap} and dataset cartography~\cite{swayamdipta-etal-2020-dataset}. 

\paragraph{Remark on diversity} In this work we are concerned about diversity at a dataset level and not an an instance level. To illustrate the difference between these two, consider the task of generating a long story. Here, it is important to ensure that the generated story has many of the features of a human written story (like complex storyline, many characters, non-repeating scenes etc.). But notice that ensuring such an instance level diversity does not guarantee diverse dataset of stories: multiple such stories obtained from an LLM could have a lot of overlap in content. For synthesis of good classification datasets, we require a more global notion of diversity which is at the dataset level. 


\section{Results}
\label{sec:expts}

\subsection{Comparison to CFG}
We compare the effect of contrast as next-token generation proceeds in CFG and \corrsyn{}. To this end, we consider \IMDb{}, and sample continuations for five 3-shot prompts from both CFG and \corrsyn{} for the same Cross-label parameters: \mbox{$\{R=1, \gamma=1.0, \delta=0.9, \alpha=0\}$}. For each token, we store the maximum absolute difference of the current label logits vector and the contrast label logits vector (i.e. $\infty$-norm of logits difference of numerator and denominator in \eqref{eq:2corr_eq1} for \corrsyn{}, and similar terms in CFG). We plot this difference against the tokens generated. 

\autoref{fig:cfg_vs_corrsynth} shows the difference between CFG and \corrsyn{}: as token generation proceeds, the effect of contrast in CFG is muted. This happens since the same generated sequence for the current label is fed back into the contrast model and thus the effect of the contrastive prompt reduces over later token generations. Whereas in \corrsyn{}, the effect of the guidance or contrast persists. As a result, we believe \corrsyn{} is a better suited for longer generations where guidance is required for the entirety of generation. In terms of complexity, as discussed previously, we incur a much higher complexity of LLM model forward passes in CFG (detailed comparison in \aref{app:compute_complexity}).

\subsection{Comparison to \fewgen}
\label{sec:corrsyn_results}
In this section, we present our experimental results against \fewgen. We use the following settings:

\insection[:]{\corrsyn{} Cross-label} Repeat factor $R=1$, Uniform contrastive guidance with \mbox{$\gamma=1.0 $} and $\delta=0.9\times\gamma$ and plausibility criterion $\alpha=10^{-3}$.

\insection[:]{\corrsyn{} Intra-label} Repeat factor $R=2$, Uniform contrastive guidance with \mbox{$\gamma=1.0 $} and $\delta=0.5\times\gamma$ and plausibility criterion $\alpha=10^{-3}$.

\insection[:]{\corrsyn{} Hybrid} Repeat factor $R=2$, set $\gamma=1.0$, Set $\gamma_{intra}=\gamma/2$, $\gamma_{cross}=\gamma/10$. Then uniform contrastive guidance in each of intra and cross terms. We set plausibility criterion \mbox{$\alpha=10^{-3}$}.

We observe in \autoref{tab:accuracy-diversity-icl} that 3-shot \corrsyn{} outperforms \fewgen{} on all evaluation metrics. Specifically, using Hybrid and Intra variants, we can achieve better student model accuracy (\DistilBERT) while increasing diversity (lower Self-BLEU, higher entity entropy) and better match with human-written text (better MAUVE). For MAUVE computation, we have used embeddings based on a GPT-2XL model. We have only shown the results for Intra and Hybrid variants since from our ablations they performed best. In \autoref{app:zeroshot}, we note the zero-shot results, which demonstrate comparable gains on all metrics.

\begin{figure*}[!t] 
    \centering
    \begin{subfigure}[t]{\textwidth}
        \centering
        \includegraphics[width=\textwidth]{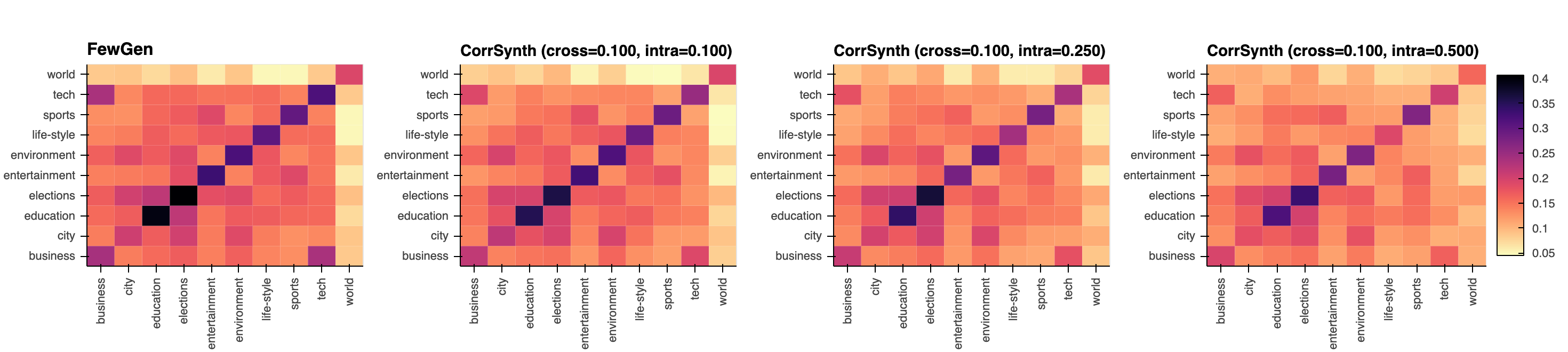}
        \caption{Impact of increasing Intra-label contrast (left to right) in Hybrid \corrsyn\ upon label-wise cosine similarities.}
        \label{fig:cosine_intra}
    \end{subfigure}
    \vskip\baselineskip
    \vspace{-0.7cm}
    \begin{subfigure}[t]{\textwidth}
        \centering
        \includegraphics[width=\textwidth]{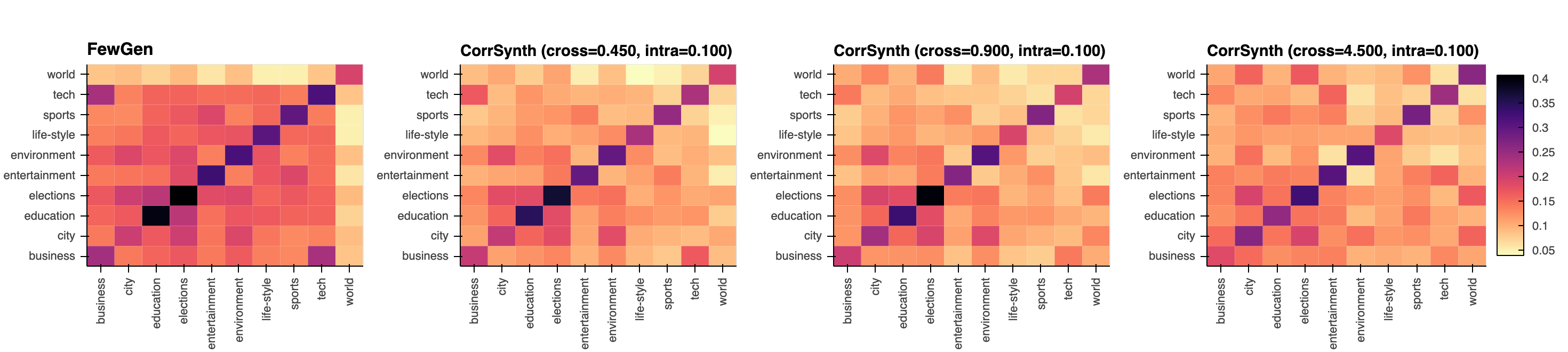}
        \caption{Impact of increasing Cross-label contrast (left to right) in Hybrid \corrsyn\ upon label-wise cosine similarities.}
        \label{fig:cosine_cross}
    \end{subfigure}
    \caption{Heatmaps for label-wise cosine similarities on \ToIHeadlines\ (with Phi-3-mini) as we increase Intra-label contrast vs increasing cross-label contrast. Note that ``Cross'' and ``Intra'' in figure titles correspond to $\gamma_{cross}$ and $\gamma_{intra}$ respectively. \fewgen\ heatmaps are provided for reference.}
    \vspace{-1em}
    \label{fig:cosine}
\end{figure*}

\subsection{Comparison to prior works} 

In \autoref{tab:baselines} we compare \corrsyn{} to current dataset generation methods as baselines. Baseline numbers are quoted from \citet{divekar2024synthesizrr}, where all results are reported on 6k rows using \DistilBERT\ student (same as our setup). The following SOTA generation methods have been compared: 
(1) \textbf{\ReGen{}} \cite{yu-etal-2023-regen}: uses dual BERT models - one for retrieval and one as a classifier - to perform multi-round filtering and eliminate noisy data based on model consistency; 
(2) \textbf{\SynthesizRR{}} \cite{divekar2024synthesizrr}: develops a hybrid retrieval-augmentation based approach to rewrite contexts, greatly enhancing the diversity of generated text;
(3) \textbf{\SunGen{}} \cite{gao2023selfguided}: employs \ZeroGen{} \cite{ye2022zerogen} to create a substantial synthetic dataset (200k rows) and then uses a bi-level optimization algorithm to assign instance-weights to each synthetic example; 
(4) \textbf{\LetsSynth{}} \cite{wang-etal-2023-lets}: builds a distinct ``seed dataset'' to train a student model, leverages an LLM to identify errors, and synthesizes supplementary data. This cycle of data augmentation is repeated.
(5) \textbf{\AttrPrompt{}} \cite{yu2023large}: enhances dataset diversity and unbiasedness by prompting a potent LLM like \ChatGPT{} with attributes identified through human-in-the-loop task analysis.

We divide our comparison into non-retrieval and retrieval based synthesis, as the latter naturally demonstrates higher diversity \citep{divekar2024synthesizrr}. We observe that \corrsyn{} achieves strong performance on all metrics, despite using a small teacher LLM (\PhiMini with 3.8B parameters) compared to prior approaches.

\begin{figure*}[!t]
\centering
    \includegraphics[width=\textwidth]{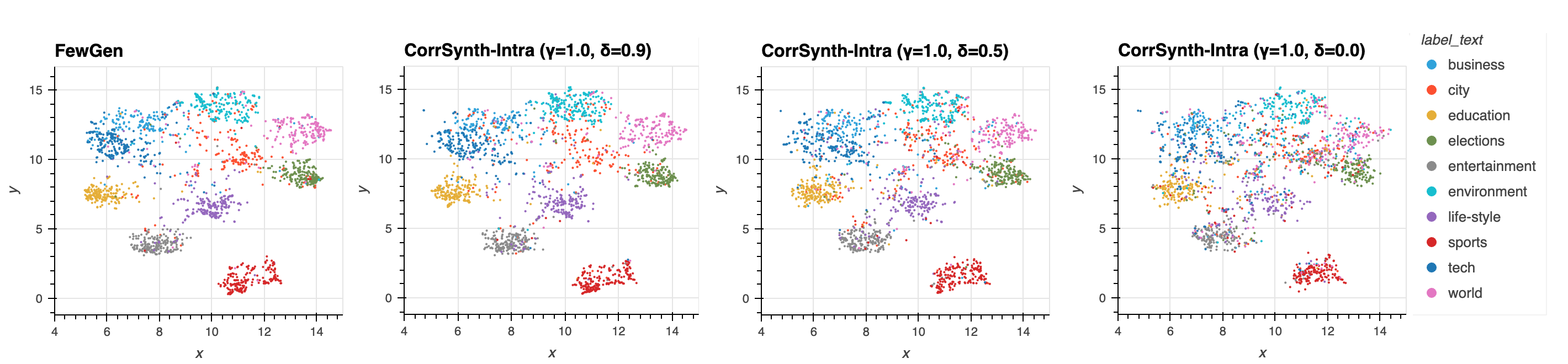}
    \vspace{-0.8cm}
    \caption{Visualising two-dimensional text representations of generations (on \ToIHeadlines\ with \PhiMini) using \corrsyn-Intra. We gradually increase guidance delta, $\delta$ in $(0.0, 0.5, 0.9)$. \fewgen\ plot is provided as a reference to the unmodified clusters (it is equivalent to $\delta=1$ i.e. no contrast).}
    \vspace{-1em}
    \label{fig:intra_label_umaps}
\end{figure*}

\begin{figure*}[!t]
\centering
\includegraphics[width=\textwidth]{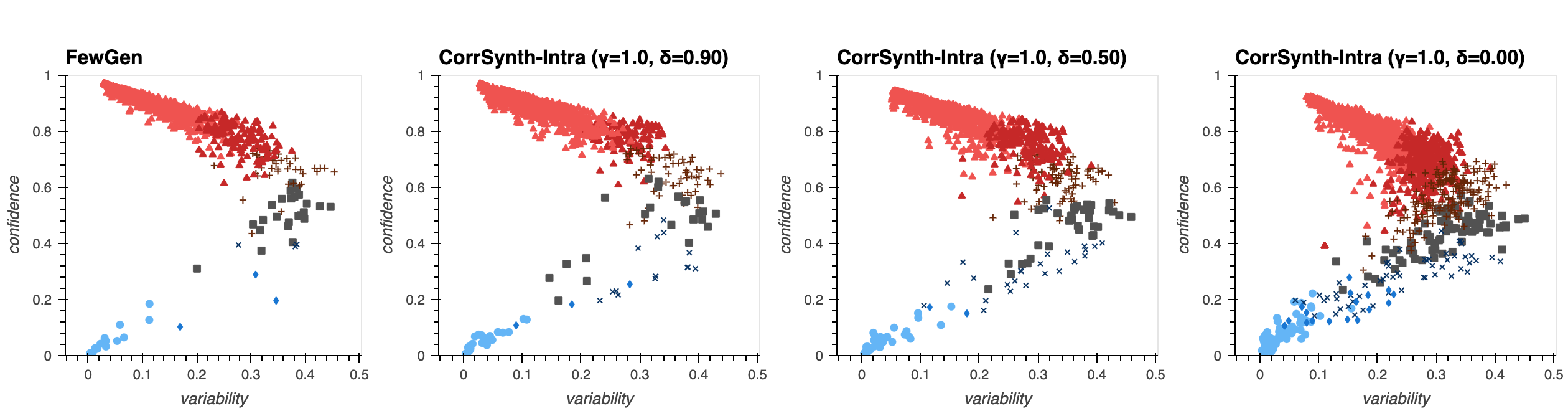}
\vspace{-0.5cm}
\caption{Datamaps for \DistilBERT\ training run on $2$K examples of \ToIHeadlines\ generated using \corrsyn-Intra using Phi-3-mini. \fewgen\ datamap is provided for reference.}
\vspace{-1em}
\label{fig:intra-label-carto}
\end{figure*}

\section{Analysis and Visualizations}
\label{sec:analysis}

\textbf{Effect of Intra-label and Cross-label contrasts:} Given the promising results of our method \corrsyn, we wanted to analyse and visualize the effect of varying Intra-label and cross-label contrasts upon the generations. For this, we obtain the average label-wise cosine similarities of the generations and plot them as heatmaps (see \autoref{fig:cosine}). We specifically study the behavior of our approach in multi-label setting upon \ToIHeadlines\ dataset to emphasize our results. In practice, we use the \texttt{all-mpnet-base-v2} model from SentenceTransformers library to obtain the text representations of each generation. Next, for generations each label $i$ and $j$ in the labelspace, we compute the pairwise cosine similarities of all generations corresponding to label $i$, to that of label $j$. The pairwise cosine similarities are then added and the mean label-wise similarity is computed. We hypothesize that in our approach (Hybrid \corrsyn), if the Intra label contrast is increased, then, within each class the generations should get further away so that the net cosine similarity within the class should reduce across all classes. As we can see in \autoref{fig:cosine_intra}, the diagonals become lighter as we go left to right. On the other hand, if the cross-label contrast is increased net separability between every pair of classes should increase. As expected, we can see in \autoref{fig:cosine_cross}, in the heatmaps the off-diagonal label similarities are becoming lighter as cross contrast is increased.

\textbf{Effect of $\delta$:} To visualize the effect of varying $\delta$ on the generations obtained through \corrsyn, we plot 2-dimensional representations of the generations. We use Uniform Manifold Approximation and Projection (UMAP)~\cite{mcinnes2020umap} for Dimension Reduction\footnote{https://umap-learn.readthedocs.io/en/latest/} of text representations obtained using \texttt{all-mpnet-base-v2} model. As earlier, we perform this analysis in a multi-label setting with \ToIHeadlines. 

In \autoref{fig:intra_label_umaps}, we can see that as $\delta$ is reduced from $(0.9, 0.5, 0)$, the representations become more and more diffused with each other, leading to overlaps, making the student model hard to learn the decision boundary. For $\delta=0.9$, we can visualize the clusters containing data points from different labels are well-separated, which resonates with our best performing results as well. Note that overlapping/diffused datapoints could be indicators of mislabelled generations as well as hard negatives. 

We hypothesize that as we decrease delta from $0.9$, first we see an increase in hard negative generations than mislabeled generations, whereas after some threshold, the extent of mislabeled generations increase. Thus there is a sweet spot which provides good amount of hard examples with minimal number of wrong generations. We can see this effect in the corresponding cartography plots \cite{swayamdipta-etal-2020-dataset} in \autoref{fig:intra-label-carto} where as we go from left to right, the density of gray and blue points increase but blue points density increases more for much smaller delta than for gray points. The gray points here typically denote hard to learn examples, where as the blue one predominantly represent mislabeled example. These hard negative generations benefit the student model training. 
\section{Related Work}
\textbf{Dataset synthesis using LLMs.} In recent years LLMs have exhibited strong generative capabilities \cite{NEURIPS2020_1457c0d6, Cobbe2021TrainingVT} to solve a diverse range of tasks. With well-designed prompts, large-scale LLMs have shown its notable zero-shot and few-shot learning ability \cite{shin-etal-2020-autoprompt, jiang-etal-2020-know, 10.1145/3411763.3451760}. 

More recently, these models have gained popularity in their superior ability to synthesize task-specific datasets \cite{Wang2021TowardsZL, Lee2021NeuralDA, kumar-etal-2020-data, puri-etal-2020-training, AnabyTavor2019DoNH}. LLMs such as GPT-3 \cite{wang-etal-2023-self-instruct, honovich-etal-2023-unnatural, west-etal-2022-symbolic} and chat-tuned models \cite{Yehudai2024GenieAH, yu2023large, wang-etal-2023-lets} have shown promising results on the task of generating synthetic data. Certain works like \citet{Meng2023TuningLM} fine-tune an LLM to generate NLU datasets, whereas our work is similar to \citet{schick-schutze-2021-generating,Meng2022GeneratingTD} which use frozen LLMs with task-dependent prompts to generate data, targeting text classification.

\textbf{Text classification dataset synthesis} employs class-specific prompts; previous studies explored zero-shot \cite{Ye2022ZeroGenEZ} and iterative few-shot prompting \cite{ye-etal-2022-progen}. However, only recently has the lack of diversity in synthetic classification datasets been recognized. \citet{yu2023large} advocated for using diverse prompts that explicitly introduce variations, such as subtopics and brands, resulting in more diverse conditioning. In contrast, our approach achieves diversity with a fixed prompt. \citet{divekar2024synthesizrr} employs retrieval augmentation to introduce variety into the dataset synthesis process by seeding the LLM with different content. However, the diversity here is constrained by the corpus availability, whereas our work improves diversity despite relying only on LLM parameters.


\textbf{Classifier-Free Guidance (CFG)} is a sampling technique introduced in diffusion literature~\cite{ho2021classifierfree} and later extended to autoregressive LLMs~\cite{sanchez2023stay}. CFG falls under general guidance based techniques, where a guidance distribution is used at inference to alter the sampling distribution towards the desired goal.In CFG, this guidance distribution is provided by the LLM itself but with a different prompt as described in \aref{sec:CFG}. Context-aware decoding~\citet{Shi2023TrustingYE} also uses the same formulation as CFG.

\textbf{Contrastive decoding (CD} refers to another family of guidance based methods that derive the guidance distribution from either a smaller LLM~\cite{o2023contrastive,li2023contrastive}, different layers of the same LLM~\cite{chuang2023dola,gera-etal-2023-benefits}. In all these methods from CFG to CD, the idea is essentially that to generate a sequence, a contrasting distribution is computed at inference. But different sequences are generated independently. In \corrsyn{}, although we borrow the general idea of a using a guidance-providing distribution at inference, the guidance distribution itself corresponds to a actual parallel generation providing both a) (anti-)correlation between multiple sequences as desired, b) compute efficiency. See \autoref{sec:method} and \autoref{sec:compare_cfg_corrsyn_app}.

\section{Conclusion}
In this work, we propose a novel technique \corrsyn\ which uses correlated sampling and intuition from classifier free guidance and contrastive decoding to generate strongly diverse datasets across a variety of tasks with good cross-label separations. We provide the mathematical intuition of our approach and back our theoretical discussion with empirical results. Our extensive experimentation across 4 datasets show the robustness of our approach in generating synthetic data. 

In the future, we would like to study the effect of including Intra-label contrasts while generating with the LLMs, and mixing up both cross-label and Intra-label contrasts (a hybrid approach) to see how the generations are affected with respect to both intrinsic and extrinsic evaluation.
\section{Limitations}
The scope of our experiments is restricted to a set of classification tasks over a few English domains of text. While we believe our approach can be applied to other languages, other domains, and tasks like question answering that go beyond classification, we have not validated this in this work. Furthermore, the scope of our formulation is restricted to supervised learning problems where there a well-defined or natural label space. Extensions to unsupervised tasks like datasets for pre-training is an interesting possibility to be explored. The introduction of new hyper-parameters in any method requires tuning, which increases costs. In our case a high value of $\delta$ with respect to the original guidance $\gamma$ (e.g. $\delta = 0.9*\gamma$, yields positive results for all guidance values). However, the tuning of the initial guidance parameter was subject to a heuristic search. Finally, our approach performs modifications to the generation process by performing correlated sampling in the logits space. This makes our approach infeasible to use with API-only teacher LMs such as GPT-4, Claude, Gemini, etc.

\bibliography{custom, refs}

\begin{thebibliography}{58}
\providecommand{\natexlab}[1]{#1}

\bibitem[{Abdin et~al.(2024)Abdin, Jacobs, Awan, Aneja, Awadallah, Awadalla,
  Bach, Bahree, Bakhtiari, Behl et~al.}]{abdin2024phi}
Marah Abdin, Sam~Ade Jacobs, Ammar~Ahmad Awan, Jyoti Aneja, Ahmed Awadallah,
  Hany Awadalla, Nguyen Bach, Amit Bahree, Arash Bakhtiari, Harkirat Behl,
  et~al. 2024.
\newblock Phi-3 technical report: A highly capable language model locally on
  your phone.
\newblock \emph{arXiv preprint arXiv:2404.14219}.

\bibitem[{Achiam et~al.(2023)Achiam, Adler, Agarwal, Ahmad, Akkaya, Aleman,
  Almeida, Altenschmidt, Altman, Anadkat, Avila, Babuschkin, Balaji, Balcom,
  Baltescu, Bao, Bavarian, Belgum, Bello, Berdine, Bernadett-Shapiro, Berner,
  Bogdonoff, Boiko, Boyd, Brakman, Brockman, Brooks, Brundage, Button, Cai,
  Campbell, Cann, Carey, Carlson, Carmichael, Chan, Chang, Chantzis, Chen,
  Chen, Chen, Chen, Chen, Chess, Cho, Chu, Chung, Cummings, Currier, Dai,
  Decareaux, Degry, Deutsch, Deville, Dhar, Dohan, Dowling, Dunning, Ecoffet,
  Eleti, Eloundou, Farhi, Fedus, Felix, Fishman, Forte, Fulford, Gao, Georges,
  Gibson, Goel, Gogineni, Goh, Gontijo-Lopes, Gordon, Grafstein, Gray, Greene,
  Gross, Gu, Guo, Hallacy, Han, Harris, He, Heaton, Heidecke, Hesse, Hickey,
  Hickey, Hoeschele, Houghton, Hsu, Hu, Hu, Huizinga, Jain, Jain, Jang, Jiang,
  Jiang, Jin, Jin, Jomoto, Jonn, Jun, Kaftan, Kaiser, Kamali, Kanitscheider,
  Keskar, Khan, Kilpatrick, Kim, Kim, Kim, Kirchner, Kiros, Knight, Kokotajlo,
  Kondraciuk, Kondrich, Konstantinidis, Kosic, Krueger, Kuo, Lampe, Lan, Lee,
  Leike, Leung, Levy, Li, Lim, Lin, Lin, Litwin, Lopez, Lowe, Lue, Makanju,
  Malfacini, Manning, Markov, Markovski, Martin, Mayer, Mayne, McGrew,
  McKinney, McLeavey, McMillan, McNeil, Medina, Mehta, Menick, Metz,
  Mishchenko, Mishkin, Monaco, Morikawa, Mossing, Mu, Murati, Murk, M'ely,
  Nair, Nakano, Nayak, Neelakantan, Ngo, Noh, Long, O'Keefe, Pachocki, Paino,
  Palermo, Pantuliano, Parascandolo, Parish, Parparita, Passos, Pavlov, Peng,
  Perelman, de~Avila Belbute~Peres, Petrov, de~Oliveira~Pinto, Pokorny,
  Pokrass, Pong, Powell, Power, Power, Proehl, Puri, Radford, Rae, Ramesh,
  Raymond, Real, Rimbach, Ross, Rotsted, Roussez, Ryder, Saltarelli, Sanders,
  Santurkar, Sastry, Schmidt, Schnurr, Schulman, Selsam, Sheppard, Sherbakov,
  Shieh, Shoker, Shyam, Sidor, Sigler, Simens, Sitkin, Slama, Sohl, Sokolowsky,
  Song, Staudacher, Such, Summers, Sutskever, Tang, Tezak, Thompson, Tillet,
  Tootoonchian, Tseng, Tuggle, Turley, Tworek, Uribe, Vallone, Vijayvergiya,
  Voss, Wainwright, Wang, Wang, Wang, Ward, Wei, Weinmann, Welihinda, Welinder,
  Weng, Weng, Wiethoff, Willner, Winter, Wolrich, Wong, Workman, Wu, Wu, Wu,
  Xiao, Xu, Yoo, Yu, Yuan, Zaremba, Zellers, Zhang, Zhang, Zhao, Zheng, Zhuang,
  Zhuk, and Zoph}]{Achiam2023GPT4TR}
OpenAI~Josh Achiam, Steven Adler, Sandhini Agarwal, Lama Ahmad, Ilge Akkaya,
  Florencia~Leoni Aleman, Diogo Almeida, Janko Altenschmidt, Sam Altman,
  Shyamal Anadkat, Red Avila, Igor Babuschkin, Suchir Balaji, Valerie Balcom,
  Paul Baltescu, Haiming Bao, Mo~Bavarian, Jeff Belgum, Irwan Bello, Jake
  Berdine, Gabriel Bernadett-Shapiro, Christopher Berner, Lenny Bogdonoff, Oleg
  Boiko, Madelaine Boyd, Anna-Luisa Brakman, Greg Brockman, Tim Brooks, Miles
  Brundage, Kevin Button, Trevor Cai, Rosie Campbell, Andrew Cann, Brittany
  Carey, Chelsea Carlson, Rory Carmichael, Brooke Chan, Che Chang, Fotis
  Chantzis, Derek Chen, Sully Chen, Ruby Chen, Jason Chen, Mark Chen, Benjamin
  Chess, Chester Cho, Casey Chu, Hyung~Won Chung, Dave Cummings, Jeremiah
  Currier, Yunxing Dai, Cory Decareaux, Thomas Degry, Noah Deutsch, Damien
  Deville, Arka Dhar, David Dohan, Steve Dowling, Sheila Dunning, Adrien
  Ecoffet, Atty Eleti, Tyna Eloundou, David Farhi, Liam Fedus, Niko Felix,
  Sim'on~Posada Fishman, Juston Forte, Isabella Fulford, Leo Gao, Elie Georges,
  Christian Gibson, Vik Goel, Tarun Gogineni, Gabriel Goh, Raphael
  Gontijo-Lopes, Jonathan Gordon, Morgan Grafstein, Scott Gray, Ryan Greene,
  Joshua Gross, Shixiang~Shane Gu, Yufei Guo, Chris Hallacy, Jesse Han, Jeff
  Harris, Yuchen He, Mike Heaton, Johannes Heidecke, Chris Hesse, Alan Hickey,
  Wade Hickey, Peter Hoeschele, Brandon Houghton, Kenny Hsu, Shengli Hu, Xin
  Hu, Joost Huizinga, Shantanu Jain, Shawn Jain, Joanne Jang, Angela Jiang,
  Roger Jiang, Haozhun Jin, Denny Jin, Shino Jomoto, Billie Jonn, Heewoo Jun,
  Tomer Kaftan, Lukasz Kaiser, Ali Kamali, Ingmar Kanitscheider, Nitish~Shirish
  Keskar, Tabarak Khan, Logan Kilpatrick, Jong~Wook Kim, Christina Kim, Yongjik
  Kim, Hendrik Kirchner, Jamie~Ryan Kiros, Matthew Knight, Daniel Kokotajlo,
  Lukasz Kondraciuk, Andrew Kondrich, Aris Konstantinidis, Kyle Kosic, Gretchen
  Krueger, Vishal Kuo, Michael Lampe, Ikai Lan, Teddy Lee, Jan Leike, Jade
  Leung, Daniel Levy, Chak~Ming Li, Rachel Lim, Molly Lin, Stephanie Lin,
  Mateusz Litwin, Theresa Lopez, Ryan Lowe, Patricia Lue, Anna~Adeola Makanju,
  Kim Malfacini, Sam Manning, Todor Markov, Yaniv Markovski, Bianca Martin,
  Katie Mayer, Andrew Mayne, Bob McGrew, Scott~Mayer McKinney, Christine
  McLeavey, Paul McMillan, Jake McNeil, David Medina, Aalok Mehta, Jacob
  Menick, Luke Metz, Andrey Mishchenko, Pamela Mishkin, Vinnie Monaco, Evan
  Morikawa, Daniel~P. Mossing, Tong Mu, Mira Murati, Oleg Murk, David M'ely,
  Ashvin Nair, Reiichiro Nakano, Rajeev Nayak, Arvind Neelakantan, Richard Ngo,
  Hyeonwoo Noh, Ouyang Long, Cullen O'Keefe, Jakub~W. Pachocki, Alex Paino, Joe
  Palermo, Ashley Pantuliano, Giambattista Parascandolo, Joel Parish, Emy
  Parparita, Alexandre Passos, Mikhail Pavlov, Andrew Peng, Adam Perelman,
  Filipe de~Avila Belbute~Peres, Michael Petrov, Henrique~Pond{\'e}
  de~Oliveira~Pinto, Michael Pokorny, Michelle Pokrass, Vitchyr~H. Pong, Tolly
  Powell, Alethea Power, Boris Power, Elizabeth Proehl, Raul Puri, Alec
  Radford, Jack Rae, Aditya Ramesh, Cameron Raymond, Francis Real, Kendra
  Rimbach, Carl Ross, Bob Rotsted, Henri Roussez, Nick Ryder, Mario~D.
  Saltarelli, Ted Sanders, Shibani Santurkar, Girish Sastry, Heather Schmidt,
  David Schnurr, John Schulman, Daniel Selsam, Kyla Sheppard, Toki Sherbakov,
  Jessica Shieh, Sarah Shoker, Pranav Shyam, Szymon Sidor, Eric Sigler, Maddie
  Simens, Jordan Sitkin, Katarina Slama, Ian Sohl, Benjamin~D. Sokolowsky, Yang
  Song, Natalie Staudacher, Felipe~Petroski Such, Natalie Summers, Ilya
  Sutskever, Jie Tang, Nikolas~A. Tezak, Madeleine Thompson, Phil Tillet, Amin
  Tootoonchian, Elizabeth Tseng, Preston Tuggle, Nick Turley, Jerry Tworek,
  Juan Felipe~Cer'on Uribe, Andrea Vallone, Arun Vijayvergiya, Chelsea Voss,
  Carroll Wainwright, Justin~Jay Wang, Alvin Wang, Ben Wang, Jonathan Ward,
  Jason Wei, CJ~Weinmann, Akila Welihinda, Peter Welinder, Jiayi Weng, Lilian
  Weng, Matt Wiethoff, Dave Willner, Clemens Winter, Samuel Wolrich, Hannah
  Wong, Lauren Workman, Sherwin Wu, Jeff Wu, Michael Wu, Kai Xiao, Tao Xu,
  Sarah Yoo, Kevin Yu, Qiming Yuan, Wojciech Zaremba, Rowan Zellers, Chong
  Zhang, Marvin Zhang, Shengjia Zhao, Tianhao Zheng, Juntang Zhuang, William
  Zhuk, and Barret Zoph. 2023.
\newblock \href {https://api.semanticscholar.org/CorpusID:257532815} {Gpt-4
  technical report}.

\bibitem[{Anaby-Tavor et~al.(2019)Anaby-Tavor, Carmeli, Goldbraich, Kantor,
  Kour, Shlomov, Tepper, and Zwerdling}]{AnabyTavor2019DoNH}
Ateret Anaby-Tavor, Boaz Carmeli, Esther Goldbraich, Amir Kantor, George Kour,
  Segev Shlomov, N.~Tepper, and Naama Zwerdling. 2019.
\newblock \href {https://api.semanticscholar.org/CorpusID:212821571} {Do not
  have enough data? deep learning to the rescue!}
\newblock In \emph{AAAI Conference on Artificial Intelligence}.

\bibitem[{Bai et~al.(2022)Bai, Jones, Ndousse, Askell, Chen, DasSarma, Drain,
  Fort, Ganguli, Henighan, Joseph, Kadavath, Kernion, Conerly, El-Showk,
  Elhage, Hatfield-Dodds, Hernandez, Hume, Johnston, Kravec, Lovitt, Nanda,
  Olsson, Amodei, Brown, Clark, McCandlish, Olah, Mann, and
  Kaplan}]{Bai2022TrainingAH}
Yuntao Bai, Andy Jones, Kamal Ndousse, Amanda Askell, Anna Chen, Nova DasSarma,
  Dawn Drain, Stanislav Fort, Deep Ganguli, Tom Henighan, Nicholas Joseph,
  Saurav Kadavath, John Kernion, Tom Conerly, Sheer El-Showk, Nelson Elhage,
  Zac Hatfield-Dodds, Danny Hernandez, Tristan Hume, Scott Johnston, Shauna
  Kravec, Liane Lovitt, Neel Nanda, Catherine Olsson, Dario Amodei, Tom~B.
  Brown, Jack Clark, Sam McCandlish, Christopher Olah, Benjamin Mann, and Jared
  Kaplan. 2022.
\newblock \href {https://api.semanticscholar.org/CorpusID:248118878} {Training
  a helpful and harmless assistant with reinforcement learning from human
  feedback}.
\newblock \emph{ArXiv}, abs/2204.05862.

\bibitem[{Brown et~al.(2020{\natexlab{a}})Brown, Mann, Ryder, Subbiah, Kaplan,
  Dhariwal, Neelakantan, Shyam, Sastry, Askell, Agarwal, Herbert-Voss, Krueger,
  Henighan, Child, Ramesh, Ziegler, Wu, Winter, Hesse, Chen, Sigler, Litwin,
  Gray, Chess, Clark, Berner, McCandlish, Radford, Sutskever, and
  Amodei}]{NEURIPS2020_1457c0d6}
Tom Brown, Benjamin Mann, Nick Ryder, Melanie Subbiah, Jared~D Kaplan, Prafulla
  Dhariwal, Arvind Neelakantan, Pranav Shyam, Girish Sastry, Amanda Askell,
  Sandhini Agarwal, Ariel Herbert-Voss, Gretchen Krueger, Tom Henighan, Rewon
  Child, Aditya Ramesh, Daniel Ziegler, Jeffrey Wu, Clemens Winter, Chris
  Hesse, Mark Chen, Eric Sigler, Mateusz Litwin, Scott Gray, Benjamin Chess,
  Jack Clark, Christopher Berner, Sam McCandlish, Alec Radford, Ilya Sutskever,
  and Dario Amodei. 2020{\natexlab{a}}.
\newblock \href
  {https://proceedings.neurips.cc/paper_files/paper/2020/file/1457c0d6bfcb4967418bfb8ac142f64a-Paper.pdf}
  {Language models are few-shot learners}.
\newblock In \emph{Advances in Neural Information Processing Systems},
  volume~33, pages 1877--1901. Curran Associates, Inc.

\bibitem[{Brown et~al.(2020{\natexlab{b}})Brown, Mann, Ryder, Subbiah, Kaplan,
  Dhariwal, Neelakantan, Shyam, Sastry, Askell, Agarwal, Herbert-Voss, Krueger,
  Henighan, Child, Ramesh, Ziegler, Wu, Winter, Hesse, Chen, Sigler, Litwin,
  Gray, Chess, Clark, Berner, McCandlish, Radford, Sutskever, and
  Amodei}]{gpt3}
Tom Brown, Benjamin Mann, Nick Ryder, Melanie Subbiah, Jared~D Kaplan, Prafulla
  Dhariwal, Arvind Neelakantan, Pranav Shyam, Girish Sastry, Amanda Askell,
  Sandhini Agarwal, Ariel Herbert-Voss, Gretchen Krueger, Tom Henighan, Rewon
  Child, Aditya Ramesh, Daniel Ziegler, Jeffrey Wu, Clemens Winter, Chris
  Hesse, Mark Chen, Eric Sigler, Mateusz Litwin, Scott Gray, Benjamin Chess,
  Jack Clark, Christopher Berner, Sam McCandlish, Alec Radford, Ilya Sutskever,
  and Dario Amodei. 2020{\natexlab{b}}.
\newblock \href
  {https://proceedings.neurips.cc/paper_files/paper/2020/file/1457c0d6bfcb4967418bfb8ac142f64a-Paper.pdf}
  {Language models are few-shot learners}.
\newblock In \emph{Advances in Neural Information Processing Systems},
  volume~33, pages 1877--1901. Curran Associates, Inc.

\bibitem[{Chang et~al.(2024)Chang, Wang, Wang, Wu, Yang, Zhu, Chen, Yi, Wang,
  Wang, Ye, Zhang, Chang, Yu, Yang, and Xie}]{llm-eval-survey}
Yupeng Chang, Xu~Wang, Jindong Wang, Yuan Wu, Linyi Yang, Kaijie Zhu, Hao Chen,
  Xiaoyuan Yi, Cunxiang Wang, Yidong Wang, Wei Ye, Yue Zhang, Yi~Chang,
  Philip~S. Yu, Qiang Yang, and Xing Xie. 2024.
\newblock \href {https://doi.org/10.1145/3641289} {A survey on evaluation of
  large language models}.
\newblock \emph{ACM Trans. Intell. Syst. Technol.}

\bibitem[{Chuang et~al.(2023)Chuang, Xie, Luo, Kim, Glass, and
  He}]{chuang2023dola}
Yung-Sung Chuang, Yujia Xie, Hongyin Luo, Yoon Kim, James Glass, and Pengcheng
  He. 2023.
\newblock Dola: Decoding by contrasting layers improves factuality in large
  language models.
\newblock \emph{arXiv preprint arXiv:2309.03883}.

\bibitem[{Cobbe et~al.(2021)Cobbe, Kosaraju, Bavarian, Chen, Jun, Kaiser,
  Plappert, Tworek, Hilton, Nakano, Hesse, and Schulman}]{Cobbe2021TrainingVT}
Karl Cobbe, Vineet Kosaraju, Mohammad Bavarian, Mark Chen, Heewoo Jun, Lukasz
  Kaiser, Matthias Plappert, Jerry Tworek, Jacob Hilton, Reiichiro Nakano,
  Christopher Hesse, and John Schulman. 2021.
\newblock \href {https://api.semanticscholar.org/CorpusID:239998651} {Training
  verifiers to solve math word problems}.
\newblock \emph{ArXiv}, abs/2110.14168.

\bibitem[{Devlin et~al.(2019)Devlin, Chang, Lee, and
  Toutanova}]{devlin-etal-2019-bert}
Jacob Devlin, Ming-Wei Chang, Kenton Lee, and Kristina Toutanova. 2019.
\newblock \href {https://doi.org/10.18653/v1/N19-1423} {{BERT}: Pre-training of
  deep bidirectional transformers for language understanding}.
\newblock In \emph{Proceedings of the 2019 Conference of the North {A}merican
  Chapter of the Association for Computational Linguistics: Human Language
  Technologies, Volume 1 (Long and Short Papers)}, pages 4171--4186,
  Minneapolis, Minnesota. Association for Computational Linguistics.

\bibitem[{Divekar and Durrett(2024)}]{divekar2024synthesizrr}
Abhishek Divekar and Greg Durrett. 2024.
\newblock Synthesizrr: Generating diverse datasets with retrieval augmentation.
\newblock \emph{arXiv preprint arXiv:2405.10040}.

\bibitem[{Gao et~al.(2022)Gao, Pi, Yong, Xu, Ye, Wu, ZHANG, Liang, Li, and
  Kong}]{gao2022self}
Jiahui Gao, Renjie Pi, LIN Yong, Hang Xu, Jiacheng Ye, Zhiyong Wu, WEIZHONG
  ZHANG, Xiaodan Liang, Zhenguo Li, and Lingpeng Kong. 2022.
\newblock {Self-Guided Noise-Free Data Generation for Efficient Zero-Shot
  Learning}.
\newblock In \emph{The Eleventh International Conference on Learning
  Representations}.

\bibitem[{Gao et~al.(2023)Gao, Pi, Yong, Xu, Ye, Wu, Zhang, Liang, Li, and
  Kong}]{gao2023selfguided}
Jiahui Gao, Renjie Pi, LIN Yong, Hang Xu, Jiacheng Ye, Zhiyong Wu, Weizhong
  Zhang, Xiaodan Liang, Zhenguo Li, and Lingpeng Kong. 2023.
\newblock \href {https://openreview.net/forum?id=h5OpjGd_lo6} {Self-guided
  noise-free data generation for efficient zero-shot learning}.
\newblock In \emph{The Eleventh International Conference on Learning
  Representations}.

\bibitem[{Gera et~al.(2023)Gera, Friedman, Arviv, Gunasekara, Sznajder, Slonim,
  and Shnarch}]{gera-etal-2023-benefits}
Ariel Gera, Roni Friedman, Ofir Arviv, Chulaka Gunasekara, Benjamin Sznajder,
  Noam Slonim, and Eyal Shnarch. 2023.
\newblock \href {https://doi.org/10.18653/v1/2023.acl-long.580} {The benefits
  of bad advice: Autocontrastive decoding across model layers}.
\newblock In \emph{Proceedings of the 61st Annual Meeting of the Association
  for Computational Linguistics (Volume 1: Long Papers)}, pages 10406--10420,
  Toronto, Canada. Association for Computational Linguistics.

\bibitem[{Guo and Chen(2024)}]{guo2024generative}
Xu~Guo and Yiqiang Chen. 2024.
\newblock {Generative AI for Synthetic Data Generation: Methods, Challenges and
  the Future}.
\newblock \emph{arXiv preprint arXiv:2403.04190}.

\bibitem[{Ho and Salimans(2021)}]{ho2021classifierfree}
Jonathan Ho and Tim Salimans. 2021.
\newblock \href {https://openreview.net/forum?id=qw8AKxfYbI} {{Classifier-Free
  Diffusion Guidance}}.
\newblock In \emph{NeurIPS 2021 Workshop on Deep Generative Models and
  Downstream Applications}.

\bibitem[{Honovich et~al.(2023)Honovich, Scialom, Levy, and
  Schick}]{honovich-etal-2023-unnatural}
Or~Honovich, Thomas Scialom, Omer Levy, and Timo Schick. 2023.
\newblock \href {https://doi.org/10.18653/v1/2023.acl-long.806} {Unnatural
  instructions: Tuning language models with (almost) no human labor}.
\newblock In \emph{Proceedings of the 61st Annual Meeting of the Association
  for Computational Linguistics (Volume 1: Long Papers)}, pages 14409--14428,
  Toronto, Canada. Association for Computational Linguistics.

\bibitem[{Jiang et~al.(2023)Jiang, Sablayrolles, Mensch, Bamford, Chaplot,
  Casas, Bressand, Lengyel, Lample, Saulnier et~al.}]{jiang2023mistral}
Albert~Q Jiang, Alexandre Sablayrolles, Arthur Mensch, Chris Bamford,
  Devendra~Singh Chaplot, Diego de~las Casas, Florian Bressand, Gianna Lengyel,
  Guillaume Lample, Lucile Saulnier, et~al. 2023.
\newblock Mistral 7b.
\newblock \emph{arXiv preprint arXiv:2310.06825}.

\bibitem[{Jiang et~al.(2024)Jiang, Sablayrolles, Roux, Mensch, Savary, Bamford,
  Chaplot, Casas, Hanna, Bressand et~al.}]{jiang2024mixtral}
Albert~Q Jiang, Alexandre Sablayrolles, Antoine Roux, Arthur Mensch, Blanche
  Savary, Chris Bamford, Devendra~Singh Chaplot, Diego de~las Casas, Emma~Bou
  Hanna, Florian Bressand, et~al. 2024.
\newblock Mixtral of experts.
\newblock \emph{arXiv preprint arXiv:2401.04088}.

\bibitem[{Jiang et~al.(2020)Jiang, Xu, Araki, and
  Neubig}]{jiang-etal-2020-know}
Zhengbao Jiang, Frank~F. Xu, Jun Araki, and Graham Neubig. 2020.
\newblock \href {https://doi.org/10.1162/tacl_a_00324} {How can we know what
  language models know?}
\newblock \emph{Transactions of the Association for Computational Linguistics},
  8:423--438.

\bibitem[{Kulkarni(2020)}]{toiheadlines}
Rohit Kulkarni. 2020.
\newblock \href {https://doi.org/10.7910/DVN/DPQMQH} {{Times of India News
  Headlines}}.

\bibitem[{Kumar et~al.(2020)Kumar, Choudhary, and Cho}]{kumar-etal-2020-data}
Varun Kumar, Ashutosh Choudhary, and Eunah Cho. 2020.
\newblock \href {https://aclanthology.org/2020.lifelongnlp-1.3} {Data
  augmentation using pre-trained transformer models}.
\newblock In \emph{Proceedings of the 2nd Workshop on Life-long Learning for
  Spoken Language Systems}, pages 18--26, Suzhou, China. Association for
  Computational Linguistics.

\bibitem[{Lee et~al.(2021)Lee, Guu, He, Dozat, and Chung}]{Lee2021NeuralDA}
Kenton Lee, Kelvin Guu, Luheng He, Timothy Dozat, and Hyung~Won Chung. 2021.
\newblock \href {https://api.semanticscholar.org/CorpusID:231749880} {{Neural
  Data Augmentation via Example Extrapolation}}.
\newblock \emph{ArXiv}, abs/2102.01335.

\bibitem[{Lewis et~al.(2020)Lewis, Perez, Piktus, Petroni, Karpukhin, Goyal,
  K{\"u}ttler, Lewis, Yih, Rockt{\"a}schel et~al.}]{lewis2020retrieval}
Patrick Lewis, Ethan Perez, Aleksandra Piktus, Fabio Petroni, Vladimir
  Karpukhin, Naman Goyal, Heinrich K{\"u}ttler, Mike Lewis, Wen-tau Yih, Tim
  Rockt{\"a}schel, et~al. 2020.
\newblock Retrieval-augmented generation for knowledge-intensive nlp tasks.
\newblock \emph{Advances in Neural Information Processing Systems},
  33:9459--9474.

\bibitem[{Li et~al.(2023)Li, Holtzman, Fried, Liang, Eisner, Hashimoto,
  Zettlemoyer, and Lewis}]{li2023contrastive}
Xiang~Lisa Li, Ari Holtzman, Daniel Fried, Percy Liang, Jason Eisner,
  Tatsunori~B Hashimoto, Luke Zettlemoyer, and Mike Lewis. 2023.
\newblock {Contrastive Decoding: Open-ended Text Generation as Optimization}.
\newblock In \emph{Proceedings of the 61st Annual Meeting of the Association
  for Computational Linguistics (Volume 1: Long Papers)}, pages 12286--12312.

\bibitem[{Liu et~al.(2021)Liu, Pillutla, Welleck, Oh, Choi, and
  Harchaoui}]{liu-etal:divergence:neurips2021}
Lang Liu, Krishna Pillutla, Sean Welleck, Sewoong Oh, Yejin Choi, and Zaid
  Harchaoui. 2021.
\newblock {Divergence Frontiers for Generative Models: Sample Complexity,
  Quantization Effects, and Frontier Integrals}.
\newblock In \emph{Advances in Neural Information Processing Systems}.

\bibitem[{Maas et~al.(2011)Maas, Daly, Pham, Huang, Ng, and
  Potts}]{maas-etal-2011-learning}
Andrew~L. Maas, Raymond~E. Daly, Peter~T. Pham, Dan Huang, Andrew~Y. Ng, and
  Christopher Potts. 2011.
\newblock \href {P11-1015} {Learning word vectors for sentiment analysis}.
\newblock pages 142--150, Portland, Oregon, USA.

\bibitem[{McInnes et~al.(2020)McInnes, Healy, and Melville}]{mcinnes2020umap}
Leland McInnes, John Healy, and James Melville. 2020.
\newblock \href {https://arxiv.org/abs/1802.03426} {Umap: Uniform manifold
  approximation and projection for dimension reduction}.
\newblock \emph{Preprint}, arXiv:1802.03426.

\bibitem[{Meng et~al.(2022{\natexlab{a}})Meng, Huang, Zhang, and
  Han}]{Meng2022GeneratingTD}
Yu~Meng, Jiaxin Huang, Yu~Zhang, and Jiawei Han. 2022{\natexlab{a}}.
\newblock \href {https://api.semanticscholar.org/CorpusID:246680398}
  {Generating training data with language models: Towards zero-shot language
  understanding}.
\newblock \emph{ArXiv}, abs/2202.04538.

\bibitem[{Meng et~al.(2022{\natexlab{b}})Meng, Huang, Zhang, and
  Han}]{meng_supergen}
Yu~Meng, Jiaxin Huang, Yu~Zhang, and Jiawei Han. 2022{\natexlab{b}}.
\newblock \href
  {https://proceedings.neurips.cc/paper_files/paper/2022/file/0346c148ba1c21c6b4780a961ea141dc-Paper-Conference.pdf}
  {Generating training data with language models: Towards zero-shot language
  understanding}.
\newblock In \emph{Advances in Neural Information Processing Systems},
  volume~35, pages 462--477. Curran Associates, Inc.

\bibitem[{Meng et~al.(2023{\natexlab{a}})Meng, Michalski, Huang, Zhang,
  Abdelzaher, and Han}]{meng2023tuning}
Yu~Meng, Martin Michalski, Jiaxin Huang, Yu~Zhang, Tarek Abdelzaher, and Jiawei
  Han. 2023{\natexlab{a}}.
\newblock Tuning language models as training data generators for
  augmentation-enhanced few-shot learning.
\newblock In \emph{International Conference on Machine Learning}, pages
  24457--24477. PMLR.

\bibitem[{Meng et~al.(2023{\natexlab{b}})Meng, Michalski, Huang, Zhang,
  Abdelzaher, and Han}]{Meng2023TuningLM}
Yu~Meng, Martin Michalski, Jiaxin Huang, Yu~Zhang, Tarek Abdelzaher, and Jiawei
  Han. 2023{\natexlab{b}}.
\newblock Tuning language models as training data generators for
  augmentation-enhanced few-shot learning.
\newblock In \emph{International Conference on Machine Learning}.

\bibitem[{O'Brien and Lewis(2023)}]{o2023contrastive}
Sean O'Brien and Mike Lewis. 2023.
\newblock Contrastive decoding improves reasoning in large language models.
\newblock \emph{arXiv preprint arXiv:2309.09117}.

\bibitem[{Papineni et~al.(2002)Papineni, Roukos, Ward, and Zhu}]{bleu}
Kishore Papineni, Salim Roukos, Todd Ward, and Wei-Jing Zhu. 2002.
\newblock \href {https://doi.org/10.3115/1073083.1073135} {{BLEU: A method for
  automatic evaluation of machine translation}}.
\newblock In \emph{Proceedings of the 40th Annual Meeting on Association for
  Computational Linguistics}, ACL '02, page 311–318, USA. Association for
  Computational Linguistics.

\bibitem[{Puri et~al.(2020)Puri, Spring, Shoeybi, Patwary, and
  Catanzaro}]{puri-etal-2020-training}
Raul Puri, Ryan Spring, Mohammad Shoeybi, Mostofa Patwary, and Bryan Catanzaro.
  2020.
\newblock \href {https://doi.org/10.18653/v1/2020.emnlp-main.468} {Training
  question answering models from synthetic data}.
\newblock In \emph{Proceedings of the 2020 Conference on Empirical Methods in
  Natural Language Processing (EMNLP)}, pages 5811--5826, Online. Association
  for Computational Linguistics.

\bibitem[{Reynolds and McDonell(2021)}]{10.1145/3411763.3451760}
Laria Reynolds and Kyle McDonell. 2021.
\newblock \href {https://doi.org/10.1145/3411763.3451760} {Prompt programming
  for large language models: Beyond the few-shot paradigm}.
\newblock In \emph{Extended Abstracts of the 2021 CHI Conference on Human
  Factors in Computing Systems}, CHI EA '21, New York, NY, USA. Association for
  Computing Machinery.

\bibitem[{Sanchez et~al.(2023)Sanchez, Fan, Spangher, Levi, Ammanamanchi, and
  Biderman}]{sanchez2023stay}
Guillaume Sanchez, Honglu Fan, Alexander Spangher, Elad Levi, Pawan~Sasanka
  Ammanamanchi, and Stella Biderman. 2023.
\newblock {Stay on topic with classifier-free guidance}.
\newblock \emph{arXiv preprint arXiv:2306.17806}.

\bibitem[{Sanh et~al.(2019)Sanh, Debut, Chaumond, and
  Wolf}]{Sanh2019DistilBERT}
Victor Sanh, Lysandre Debut, Julien Chaumond, and Thomas Wolf. 2019.
\newblock \href {https://arxiv.org/abs/1910.01108} {{DistilBERT, a distilled
  version of BERT: smaller, faster, cheaper and lighter}}.
\newblock In \emph{5th Workshop on Energy Efficient Machine Learning and
  Cognitive Computing @ NeurIPS 2019}.

\bibitem[{Schick and Sch{\"u}tze(2021)}]{schick-schutze-2021-generating}
Timo Schick and Hinrich Sch{\"u}tze. 2021.
\newblock \href {https://doi.org/10.18653/v1/2021.emnlp-main.555} {Generating
  datasets with pretrained language models}.
\newblock In \emph{Proceedings of the 2021 Conference on Empirical Methods in
  Natural Language Processing}, pages 6943--6951, Online and Punta Cana,
  Dominican Republic. Association for Computational Linguistics.

\bibitem[{Shi et~al.(2023)Shi, Han, Lewis, Tsvetkov, Zettlemoyer, and
  Yih}]{Shi2023TrustingYE}
Weijia Shi, Xiaochuang Han, Mike Lewis, Yulia Tsvetkov, Luke Zettlemoyer, and
  Scott Yih. 2023.
\newblock \href {https://api.semanticscholar.org/CorpusID:258866080} {Trusting
  your evidence: Hallucinate less with context-aware decoding}.
\newblock \emph{ArXiv}, abs/2305.14739.

\bibitem[{Shin et~al.(2020)Shin, Razeghi, Logan~IV, Wallace, and
  Singh}]{shin-etal-2020-autoprompt}
Taylor Shin, Yasaman Razeghi, Robert~L. Logan~IV, Eric Wallace, and Sameer
  Singh. 2020.
\newblock \href {https://doi.org/10.18653/v1/2020.emnlp-main.346}
  {{A}uto{P}rompt: {E}liciting {K}nowledge from {L}anguage {M}odels with
  {A}utomatically {G}enerated {P}rompts}.
\newblock In \emph{Proceedings of the 2020 Conference on Empirical Methods in
  Natural Language Processing (EMNLP)}, pages 4222--4235, Online. Association
  for Computational Linguistics.

\bibitem[{Swayamdipta et~al.(2020)Swayamdipta, Schwartz, Lourie, Wang,
  Hajishirzi, Smith, and Choi}]{swayamdipta-etal-2020-dataset}
Swabha Swayamdipta, Roy Schwartz, Nicholas Lourie, Yizhong Wang, Hannaneh
  Hajishirzi, Noah~A. Smith, and Yejin Choi. 2020.
\newblock \href {https://doi.org/10.18653/v1/2020.emnlp-main.746} {Dataset
  cartography: Mapping and diagnosing datasets with training dynamics}.
\newblock In \emph{Proceedings of the 2020 Conference on Empirical Methods in
  Natural Language Processing (EMNLP)}, pages 9275--9293, Online. Association
  for Computational Linguistics.

\bibitem[{Wang et~al.(2023{\natexlab{a}})Wang, Zhou, and
  Sachan}]{wang-etal-2023-lets}
Ruida Wang, Wangchunshu Zhou, and Mrinmaya Sachan. 2023{\natexlab{a}}.
\newblock \href {https://doi.org/10.18653/v1/2023.findings-emnlp.791} {Let{'}s
  synthesize step by step: Iterative dataset synthesis with large language
  models by extrapolating errors from small models}.
\newblock In \emph{Findings of the Association for Computational Linguistics:
  EMNLP 2023}, pages 11817--11831, Singapore. Association for Computational
  Linguistics.

\bibitem[{Wang et~al.(2023{\natexlab{b}})Wang, Kordi, Mishra, Liu, Smith,
  Khashabi, and Hajishirzi}]{wang-etal-2023-self-instruct}
Yizhong Wang, Yeganeh Kordi, Swaroop Mishra, Alisa Liu, Noah~A. Smith, Daniel
  Khashabi, and Hannaneh Hajishirzi. 2023{\natexlab{b}}.
\newblock \href {https://doi.org/10.18653/v1/2023.acl-long.754} {Self-instruct:
  Aligning language models with self-generated instructions}.
\newblock In \emph{Proceedings of the 61st Annual Meeting of the Association
  for Computational Linguistics (Volume 1: Long Papers)}, pages 13484--13508,
  Toronto, Canada. Association for Computational Linguistics.

\bibitem[{Wang et~al.(2021)Wang, Yu, Firat, and Cao}]{Wang2021TowardsZL}
Zirui Wang, Adams~Wei Yu, Orhan Firat, and Yuan Cao. 2021.
\newblock \href {https://api.semanticscholar.org/CorpusID:237572306} {Towards
  zero-label language learning}.
\newblock \emph{ArXiv}, abs/2109.09193.

\bibitem[{West et~al.(2022)West, Bhagavatula, Hessel, Hwang, Jiang, Le~Bras,
  Lu, Welleck, and Choi}]{west-etal-2022-symbolic}
Peter West, Chandra Bhagavatula, Jack Hessel, Jena Hwang, Liwei Jiang, Ronan
  Le~Bras, Ximing Lu, Sean Welleck, and Yejin Choi. 2022.
\newblock \href {https://doi.org/10.18653/v1/2022.naacl-main.341} {Symbolic
  knowledge distillation: from general language models to commonsense models}.
\newblock In \emph{Proceedings of the 2022 Conference of the North American
  Chapter of the Association for Computational Linguistics: Human Language
  Technologies}, pages 4602--4625, Seattle, United States. Association for
  Computational Linguistics.

\bibitem[{Ye et~al.(2022{\natexlab{a}})Ye, Gao, Li, Xu, Feng, Wu, Yu, and
  Kong}]{ye2022zerogen}
Jiacheng Ye, Jiahui Gao, Qintong Li, Hang Xu, Jiangtao Feng, Zhiyong Wu, Tao
  Yu, and Lingpeng Kong. 2022{\natexlab{a}}.
\newblock {ZeroGen: Efficient Zero-shot Learning via Dataset Generation}.
\newblock In \emph{Proceedings of the 2022 Conference on Empirical Methods in
  Natural Language Processing}, pages 11653--11669.

\bibitem[{Ye et~al.(2022{\natexlab{b}})Ye, Gao, Li, Xu, Feng, Wu, Yu, and
  Kong}]{Ye2022ZeroGenEZ}
Jiacheng Ye, Jiahui Gao, Qintong Li, Hang Xu, Jiangtao Feng, Zhiyong Wu, Tao
  Yu, and Lingpeng Kong. 2022{\natexlab{b}}.
\newblock \href {https://api.semanticscholar.org/CorpusID:246867045} {Zerogen:
  Efficient zero-shot learning via dataset generation}.
\newblock \emph{ArXiv}, abs/2202.07922.

\bibitem[{Ye et~al.(2022{\natexlab{c}})Ye, Gao, Wu, Feng, Yu, and
  Kong}]{ye2022progen}
Jiacheng Ye, Jiahui Gao, Zhiyong Wu, Jiangtao Feng, Tao Yu, and Lingpeng Kong.
  2022{\natexlab{c}}.
\newblock {ProGen: Progressive Zero-shot Dataset Generation via In-context
  Feedback}.
\newblock In \emph{Findings of the Association for Computational Linguistics:
  EMNLP 2022}, pages 3671--3683.

\bibitem[{Ye et~al.(2022{\natexlab{d}})Ye, Gao, Wu, Feng, Yu, and
  Kong}]{ye-etal-2022-progen}
Jiacheng Ye, Jiahui Gao, Zhiyong Wu, Jiangtao Feng, Tao Yu, and Lingpeng Kong.
  2022{\natexlab{d}}.
\newblock \href {https://doi.org/10.18653/v1/2022.findings-emnlp.269}
  {{P}ro{G}en: Progressive zero-shot dataset generation via in-context
  feedback}.
\newblock In \emph{Findings of the Association for Computational Linguistics:
  EMNLP 2022}, pages 3671--3683, Abu Dhabi, United Arab Emirates. Association
  for Computational Linguistics.

\bibitem[{Yehudai et~al.(2024)Yehudai, Carmeli, Mass, Arviv, Mills, Toledo,
  Shnarch, and Choshen}]{Yehudai2024GenieAH}
Asaf Yehudai, Boaz Carmeli, Yosi Mass, Ofir Arviv, Nathaniel Mills, Assaf
  Toledo, Eyal Shnarch, and Leshem Choshen. 2024.
\newblock \href {https://api.semanticscholar.org/CorpusID:267211959} {Genie:
  Achieving human parity in content-grounded datasets generation}.
\newblock \emph{ArXiv}, abs/2401.14367.

\bibitem[{Yu et~al.(2023{\natexlab{a}})Yu, Zhuang, Zhang, Meng, Ratner,
  Krishna, Shen, and Zhang}]{yu2023large}
Yue Yu, Yuchen Zhuang, Jieyu Zhang, Yu~Meng, Alexander Ratner, Ranjay Krishna,
  Jiaming Shen, and Chao Zhang. 2023{\natexlab{a}}.
\newblock \href {https://openreview.net/forum?id=6hZIfAY9GD} {Large language
  model as attributed training data generator: A tale of diversity and bias}.
\newblock In \emph{Thirty-seventh Conference on Neural Information Processing
  Systems Datasets and Benchmarks Track}.

\bibitem[{Yu et~al.(2024)Yu, Zhuang, Zhang, Meng, Ratner, Krishna, Shen, and
  Zhang}]{yu2024large}
Yue Yu, Yuchen Zhuang, Jieyu Zhang, Yu~Meng, Alexander~J Ratner, Ranjay
  Krishna, Jiaming Shen, and Chao Zhang. 2024.
\newblock Large language model as attributed training data generator: A tale of
  diversity and bias.
\newblock \emph{Advances in Neural Information Processing Systems}, 36.

\bibitem[{Yu et~al.(2023{\natexlab{b}})Yu, Zhuang, Zhang, Meng, Shen, and
  Zhang}]{yu2023regen}
Yue Yu, Yuchen Zhuang, Rongzhi Zhang, Yu~Meng, Jiaming Shen, and Chao Zhang.
  2023{\natexlab{b}}.
\newblock Regen: Zero-shot text classification via training data generation
  with progressive dense retrieval.
\newblock \emph{arXiv preprint arXiv:2305.10703}.

\bibitem[{Yu et~al.(2023{\natexlab{c}})Yu, Zhuang, Zhang, Meng, Shen, and
  Zhang}]{yu-etal-2023-regen}
Yue Yu, Yuchen Zhuang, Rongzhi Zhang, Yu~Meng, Jiaming Shen, and Chao Zhang.
  2023{\natexlab{c}}.
\newblock \href {https://doi.org/10.18653/v1/2023.findings-acl.748} {{R}e{G}en:
  Zero-shot text classification via training data generation with progressive
  dense retrieval}.
\newblock In \emph{Findings of the Association for Computational Linguistics:
  ACL 2023}, pages 11782--11805, Toronto, Canada. Association for Computational
  Linguistics.

\bibitem[{Zhang et~al.(2015)Zhang, Zhao, and LeCun}]{zhang2015character}
Xiang Zhang, Junbo Zhao, and Yann LeCun. 2015.
\newblock Character-level convolutional networks for text classification.
\newblock In \emph{Proceedings of the 28th International Conference on Neural
  Information Processing Systems - Volume 1}, NIPS'15, page 649–657,
  Cambridge, MA, USA. MIT Press.

\bibitem[{Zhu et~al.(2018)Zhu, Lu, Zheng, Guo, Zhang, Wang, and
  Yu}]{zhu2018texygen}
Yaoming Zhu, Sidi Lu, Lei Zheng, Jiaxian Guo, Weinan Zhang, Jun Wang, and Yong
  Yu. 2018.
\newblock Texygen: A benchmarking platform for text generation models.
\newblock \emph{SIGIR}.

\bibitem[{Ziser et~al.(2020)Ziser, Kravi, and Carmel}]{humor}
Yftah Ziser, Elad Kravi, and David Carmel. 2020.
\newblock \href {https://doi.org/10.1145/3397271.3401077} {Humor detection in
  product question answering systems}.
\newblock In \emph{Proceedings of the 43rd International ACM SIGIR Conference
  on Research and Development in Information Retrieval}, SIGIR '20, page
  519–528, New York, NY, USA. Association for Computing Machinery.

\end{thebibliography}

\appendix
\section{Risks}

Although the main goal of our work is to improve text classification, our use of LLMs to generate examples does carry some conceptual risks. By generating news headlines and reviews to train classifiers on, we run the risk of generating fake news and other harmful content. However, we believe this risk is mitigated by the fact that the final outcome of our system is a classifier: classification models have relatively constrained failure modes (misclassification) compared to text generation models that can mislead users. Furthermore, we do not believe our approach uniquely advances the generation of content like fake news or reviews; our advances are largely orthogonal to the technology that brings such risks.

\section{Ablation: without in-context learning}
\label{app:zeroshot}

We explore the performance from \fewgen{} and \corrsyn{} in the absence of in-context examples. Recall that in \autoref{tab:accuracy-diversity-icl}, we used 3 in-context examples selected at random from a small seed set of 50 per class (for multiclass tasks) and 100 per class (for binary tasks). 

In this ablation, we remove this dependence completely and do not pass any in-context examples; thus, the next-token distribution is the same for each batch of contrasting terms we generate, and the variation in generations is solely a function of the top-p sampling, rather than a change to the next-token distribution which was induced due to in-context examples in the prompt.

In \autoref{tab:accuracy-diversity-zeroshot}, we observe that once again, \corrsyn{} consistently demonstrates superior diversity and accuracy compared to \fewgen. However, we note that in-context examples do improve all metrics, and thus we recommend including them in the base prompt.
\begin{table*}[h]
\centering
\begin{tabular}{
L{65pt}
c 
C{16pt} 
C{17pt} 
C{18pt} 
C{24pt} 
C{20pt} 
|C{16pt} 
C{17pt} 
C{18pt} 
C{24pt} 
C{20pt} 
}

\toprule
\multirow{2}{*}{\textbf{Method}}   
& \multirow{2}{*}{\textbf{Teacher}} 
& \multicolumn{4}{c}{\textbf{Accuracy \higherbetter}}
& \multirow{2}{*}{\textbf{Avg.}}
& \multicolumn{4}{c}{\textbf{MAUVE \higherbetter}}
& \multirow{2}{*}{\textbf{Avg.}}
\\ 
\cmidrule(l){3-6}         
\cmidrule(l){8-11}         
& \textbf{LM} 
& \textbf{\AG} 
& \textbf{\ToI} 
& \textbf{\Hum} 
& \textbf{\IMDb} 
&
& \textbf{\AG} 
& \textbf{\ToI} 
& \textbf{\Hum} 
& \textbf{\IMDb} 
&
\\ 
\midrule
\gold          
& \multicolumn{1}{c}{-}                
& 91.4         & 78.9         & 92.9          & 91.4 & 88.7
& -         & -         & -          & - & -
\\ 
\midrule
\multicolumn{12}{c}{\underline{\textsc{Zero-shot}}} \\
[0.5ex]
\fewgen 
& \multicolumn{1}{c}{\PhiMini}
& 70.3         & 53.4         & \textbf{69.0}          & 71.9 & 66.2
& 55.9         & 51.2         & 56.4          & 52.7 & 54.1
\\ 
\fewgen 
& \multicolumn{1}{c}{\Mixtral}          
& 74.0         & 51.1         & 49.1          & 64.3 & 58.1
& 50.6         & 50.0         & 52.4          & 54.1 & 51.8
\\ 
[1.0ex]

\corrsynreallyshort-Intra 
& \multicolumn{1}{c}{\PhiMini}
& 68.5         & 57.5         & 65.8          & 76.8 & 67.2
& \textbf{59.4}         & 53.7         & 62.0          & 58.4 & 58.4
\\ 
\corrsynreallyshort-Hybrid 
& \multicolumn{1}{c}{\PhiMini}
& \textbf{85.1}         & \textbf{59.3}         & 65.3          & 78.0 & \textbf{71.9}
& 57.8         & \textbf{56.7}        & \textbf{63.3}          & \textbf{58.5} & \textbf{59.1}
\\ 
[0.5ex]

\corrsynreallyshort-Intra 
& \multicolumn{1}{c}{\Mixtral}
& 74.4         & 54.5         & 52.2          & 78.1 & 64.8
& 53.6         & 50.8         & 52.4          & 55.7 & 53.1
\\  
\corrsynreallyshort-Hybrid 
& \multicolumn{1}{c}{\Mixtral}
& 73.8         & 55.0         & 58.6           & \textbf{78.7} & 66.5
& 54.1         & 51.2         & 52.6          & 56.7 & 53.7
\\ 
[1.0ex]
\toprule
\multirow{2}{*}{\textbf{Method}}   
& \multirow{2}{*}{\textbf{Teacher}} 
& \multicolumn{4}{c}{\textbf{Self-BLEU-5 \lowerbetter}}
& \multirow{2}{*}{\textbf{Avg.}}
& \multicolumn{4}{c}{\textbf{Entity-Entropy \higherbetter}}     
& \multirow{2}{*}{\textbf{Avg.}}
\\
\cmidrule(l){3-6}         
\cmidrule(l){8-11}         
& \textbf{LM} 
& \textbf{\AG} 
& \textbf{\ToI} 
& \textbf{\Hum} 
& \textbf{\IMDb} 
&
& \textbf{\AG} 
& \textbf{\ToI} 
& \textbf{\Hum} 
& \textbf{\IMDb} 
&
\\
\midrule
\gold          
& \multicolumn{1}{c}{-}                
& 17.1         & 7.9         & 19.8          & 27.9 & 18.2
& 6.6         & 6.1         & 5.1          & 7.5 & 6.3
\\ 
\midrule
\multicolumn{12}{c}{\underline{\textsc{Zero-shot}}} \\
[0.5ex]
\fewgen 
& \multicolumn{1}{c}{\PhiMini}
& 67.2         & 58.7         & 62.9          & 76.5 & 66.3
& 3.5         & 4.6         & 3.8          & 3.1 & 3.8
\\ 
\fewgen 
& \multicolumn{1}{c}{\Mixtral}
& 90.1         & 97.3         & 93.4          & 94.7 & 93.9
& 2.3         & 2.4         & 1.4          & 1.7 & 1.9
\\ 
[1.0ex]

\corrsynreallyshort-Intra 
& \multicolumn{1}{c}{\PhiMini}
& 34.8         & 28.8         & 33.8          & 51.0 & 37.1
& 4.9         & 4.8         & 4.5          & 4.4 & 4.6
\\ 
\corrsynreallyshort-Hybrid 
& \multicolumn{1}{c}{\PhiMini}
& \textbf{33.2}         & \textbf{27.8}         & \textbf{31.9}          & \textbf{46.6} & \textbf{34.9}
& \textbf{5.3}         & \textbf{5.1}         & \textbf{4.6}          & \textbf{4.8} & \textbf{5.0}
\\ 
[0.5ex]
\corrsynreallyshort-Intra 
& \multicolumn{1}{c}{\Mixtral}
& 78.1         & 87.3         & 76.9          & 84.7 & 81.8
& 3.1         & 3.4         & 2.5          & 2.8 & 3.0
\\  
\corrsynreallyshort-Hybrid 
& \multicolumn{1}{c}{\Mixtral}
& 77.4         & 86.0         & 75.0          & 81.3 & 79.9
& 3.3         & 3.3         & 2.7           & 3.1 & 3.1
\\ 
\bottomrule
\end{tabular}
\caption{
Evaluation of intrinsic dataset quality and \DistilBERT\ student model fine-tuned on real and synthetic datasets using zero-shot generation. We report mean accuracy numbers across 5 runs. 
}
\label{tab:accuracy-diversity-zeroshot}
\end{table*}

\section{\fewgen{}}
\label{sec:fewgen}
Let us consider the case of binary classification with labels $\{0,1\}$ and corresponding verbalization $\{\mathbf{y}_0,\mathbf{y}_1\}$. \fewgen{}~\cite{gpt3} is a standard approach to generate an instance $\mathbf{x}$ for a label $\mathbf{y}$: construct a prompt $\Prompt$ that has some description of the classification task, few ICL example generations, optional instance attributes and the choice of label $\mathbf{y}\in\{\mathbf{y}_0,\mathbf{y}_1\}$, and task the LLM to generate $x$. For brevity, we only keep the dependence of $\Prompt$ on $\mathbf{y}$ and use the notation $\Prompt(\mathbf{y})$ to denote the \textit{prompt tokens}. Let $\mP$ denote the auto-regressive LLM probability distribution with vocabulary $\cV$. An instance corresponding to label $\mathbf{y}$ is sampled in \fewgen{} as 
\begin{equation}
\label{eq:std_sample}
    \mathbf{x}=(x_1,\cdots,x_n)\distas{}\mP(\cdot|\Prompt(\mathbf{y}))
\end{equation}

\section{CFG}
\label{sec:CFG}
In CFG decoding~\cite{sanchez2023stay}, output token distribution is tilted in order to ensure that the LLM generations satisfy a particular condition.  In particular, we construct a \textit{contrastive prompt} $\widebar{\Prompt}$, and choose a guidance strength $\gamma>0$. Then instead of \eqref{eq:std_sample}, $\mathbf{x}$ is sampled using a titled distribution $\tilde \mP$ where
\begin{align}
    \tilde \mP(\cdot)& \propto  \frac{\mP(\cdot|\Prompt(\mathbf{y}))^{\gamma+1}}{\mP(\cdot|\widebar{\Prompt})^{\gamma}}\nonumber \\
    &=\mP(\cdot|\Prompt(\mathbf{y}))\left[\frac{\mP(\cdot|\Prompt(\mathbf{y}))}{\mP(\cdot|\widebar{\Prompt})}\right]^\gamma\label{eq:2cfg_seq}
\end{align}
Suppose we choose $\widebar{\Prompt}=\Prompt(\bar{\mathbf{y}})$, the prompt corresponding to the complementary label $\bar{\mathbf{y}}$ of $\mathbf{y}$ (or it could be any other label different from $\mathbf{y}$ in case of multiclass scenario). Then in the above equation, we are up-weighing the sequences that likely under $\Prompt(\mathbf{y})$ but unlikely under $\bar{\mathbf{y}}$ using the ratio of the two probabilities. This is supposed to move the generations away from the complementary label $\bar{\mathbf{y}}$. Writing in terms of tokens, we sample the $i$-th token $x_i$ as follows
\begin{equation}
    \label{eq:2cfg-ar}
    x_i\distas{}\tilde \mP(\cdot|\mathbf{x}_{<i}) \propto \frac{\mP(\cdot|\Prompt(\mathbf{y}),\mathbf{x}_{<i})^{\gamma+1}}{\mP(\cdot|\Prompt(\mathbf{\bar{\mathbf{y}}}),\mathbf{x}_{<i})^{\gamma}}
\end{equation}

\paragraph{Drawbacks:} We find two drawbacks in CFG:
\begin{enumerate}
    \item In equation \eqref{eq:2cfg-ar}, the same $\mathbf{x}_{<i}$ is fed as a continuation from both prompts $\Prompt(y)$ and $\Prompt(\mathbf{\bar{\mathbf{y}}})$. We posit that this leads to decrease in the effect on guidance as more tokens are generated. This is because even the generation $\mathbf{x}$ is expected to be more faithful to $\Prompt(\mathbf{y})$ than to $\Prompt(\mathbf{\bar{\mathbf{y}}})$. So even though $\Prompt(\mathbf{\bar{\mathbf{y}}})$ is sort of opposite to $\Prompt(\mathbf{y})$, feeding in the generations that are faithful to the latter would move the token distributions in the denominator closer to the numerator. This is shown in \autoref{fig:cfg_vs_corrsynth}.
    \item Only a single sequence is generated at the cost of increase in number of forward passes of the model by two-fold. So a natural $K$-way extension for $K$-class classification would incur $K^2$ forward passes through the model per token for generating a single token for each of the $K$-classes. 
\end{enumerate}

\section{Geometric mean interpretation and $K$-class \corrsyn{}}
\label{sec:geometric}
To gain intuition on \corrsyn{}, we present an interpretation of it using geometric mean. We continue to use the notation from \ref{sec:M-corrsyn}. First we present the uniform contrastive guidance described briefly in the main paper.

\subsection{Uniform contrastive guidance}

\label{sec:app_unif_guidance}
We set a parameter $\delta$ that controls the total amount of contrast guidance: for each $m$, $\sum_n \gamma_{m,n}=\gamma-\delta$. 
At step $i$, let the active set $\cS_{i}=\{m\in[M]:x_{m,i-1}\neq \eos\}\}$ which captures the sequences which have not yet hit the EOS token. Let $M_{i,active}=|\cS_{i}|$ denote the number of such sequences. Then in uniform contrastive guidance we set 
$$\gamma_{m,n}=\begin{cases} \frac{\gamma-\delta}{M_{i,active}-1}&,\, m,n\in\cS_i\\
 0&,\, \mathrm{otherwise}\end{cases}$$
at stage/token $i$ (dependence of $\gamma_{m,n}$ on $i$ is suppressed). 
Thus equation \eqref{eq:M-Corr-1} becomes
\begin{align}
    x_{m,i}&\distas{} \tilde \mP_{m,i}(\cdot)\nonumber\\
    &\propto \frac{\mP(\cdot| \Prompt_m,\mbf{x}_{m,<i})^{\gamma}}{\prod_{\substack{n\in \cS_i \\n\neq  m}}\mP(\cdot| \Prompt_n,\mbf{x}_{n,<i})^{\frac{\gamma-\delta}{M_{i,active}-1}}}\label{eq:M-Corr-2}
\end{align}

\subsection{Geometric mean}
Let us assume that $S_i=[M]$ and hence $M_{i,active}=M$. Further let $\delta=0$. Recall that the geometric mean of $n$ non-negative reals $\{\alpha_1,\cdots,\alpha_n\}$ is given by
\begin{equation}
    GM(\{\alpha_i:i\in [n]\})=\left(\prod_{i=1}^n\alpha_i\right)^{\frac{1}{n}}
\end{equation}
Analogously we can define the geometric mean of $M$ probability distributions in a point-wise manner. Thus we can write \eqref{eq:M-Corr-2} as
\begin{align}
     &x_{m,i}\distas{}\tilde \mP_{m,i}(\cdot)\nonumber\\
    &\propto \frac{\mP(\cdot| \Prompt_m,\mbf{x}_{m,<i})^{\gamma}}{GM\left(\{\mP(\cdot| \Prompt_n,\mbf{x}_{n,<i}):n\in \cS_i, n\neq m\}\right)^{\gamma}}\label{eq:M-Corr-3}
\end{align}
 Thus, in \corrsyn, the contrasting guidance signal is provided by a \textit{geometric ensemble} of token distributions obtained from contrasting prompts as well as corresponding contrasting sequences. We expect that this geometric ensemble contrast, when $M\gg 2$, to average out the signal from the contrast and mitigate the issue of non alignment of words or entities between sequences. 

 \subsection{\corrsyn{} for $K$-class data generation}
 \label{sec:K-corrsyn}
 In this section we describe how \corrsyn{} is applied to generate data for $K$-class text classification problem. Recall that in $K$-class classification problem over $\cX\times\cY$ we have classes $[K]$ with label verbalizations $\{\mbf{y}_1,\cdots,\mbf{y}_K\}$. To generates instances for each class, we create prompts as follows. Let $R\in\mathbb{N}$ be the repeat factor. For each class $\mbf{y}$ consider the, possibly empty, ICL examples sets $\cI_{\mbf{y},r}\subset \cX\times\cY$ for $r\in [R]$ which contain positive examples for $\mbf{y}$. We construct a set of $K\cdot R$ prompts $\{\Prompt_{k,r}:k\in[K],r\in[R]\}$ where $\Prompt_{k,r}=\Prompt(\mbf{y}_k,\cI_{\mbf{y}_k,r})$ is a prompt that asks the LLM to generate instances for the class ${\mbf{y_k}}$ and includes ICL examples in $\cI_{\mbf{y}_k,r}$. For brevity, we assume that no sequence hits $\eos$ until some pre-set max number of tokens has been reached. There are a couple of ways in which \corrsyn{} can be used. Here we describe just one of the ways.

 \subsubsection{Cross-label \corrsyn{}}
Here we contrast the instance for a label $\mbf{y_k}$ with instances of all the other labels $\mbf{y}_{k'}$ where $k'\neq k$. Thus, assuming uniform contrastive guidance~\ref{sec:unif_guidance}, we generate instances $\{\mbf{x}_{k,r}:k\in[K], r\in[R]\}$ together in \textit{lockstep} as follows. At stage/token $i$ we have for every $k\in[k]$ and $r\in [R]$
 \begin{align}
 \begin{aligned}
      & x_{k,r,i}\distas{} \tilde \mP_{k,r,i}(\cdot) \\
      &\propto \frac{\mP(\cdot| \Prompt_{k,r},\mbf{x}_{k,r,<i})^{\gamma}}{GM\left(\left\{\mP(\cdot| \Prompt_{k',r'},\mbf{x}_{k',r',<i})\right\}_{\substack {k'\neq k\\ r'\in[R]}}\right)^{\gamma-\delta}}\label{eq:K-Corr-cross-1}
 \end{aligned}
 \end{align}
 \paragraph{Effect of repeat factor:} We include repeat factor because it will increase the number of contrast terms for taking the geometric mean. We expect that this would provide improved averaging and reduces the noise due to potential misalignment.

 \subsubsection{Hybrid \corrsyn{}}

 In the hybrid approach, we contrast the instance $\mbf{x}_{k,r}$ for a label $\mbf{y}_k$ with instances $\mbf{x}_{k,r'}$ of the same label (but with different repeat $r'\neq r$), as well as instances $\mbf{x}_{k',r'}$ for all the other labels (where $k'\neq k$, and $r'\in [R]$). We separately set the target guidance for each of the cross and intra label terms. That is, we fix two targets $\gamma_{intra}$ and $\gamma_{cross}$. Within each group we use uniform contrastive guidance from \ref{sec:unif_guidance}. The instances are generated as follows. At stage/token $i$ we have for every $k\in[k]$ and $r\in [R]$
 
 \begin{align}
 \begin{aligned}
      x_{k,r,i}&\distas{}\tilde \mP_{k,r,i}(\cdot) \\
      &\propto \frac{\mP(\cdot| \Prompt_{k,r},\mbf{x}_{k,r,<i})^{\gamma}}{GM_{intra}^{\gamma_{intra}}\cdot GM_{cross}^{\gamma_{cross}}}\label{eq:K-Corr-hybrid-1}
 \end{aligned}
 \end{align}

 where 
 \begin{align}
 \begin{aligned}
     &GM_{intra}=\\
     &GM\left(\left\{\mP(\cdot| \Prompt_{k,r'},\mbf{x}_{k,r',<i})\right\}_{\substack {r'\neq r}}\right)\\
     &GM_{cross}=\\
     &GM\left(\left\{\mP(\cdot| \Prompt_{k',r'},\mbf{x}_{k',r',<i})\right\}_{\substack {k'\neq k\\ r'\in[R]}}\right)
\end{aligned}
 \end{align}
 
As seen from the above display, the first term in the denominator gives contrast signal from generations with the class, in order to get good intra-label diversity. While the second term gives contrast signal from other classes and hence serves to increase class separation. 

\subsection{\corrsyn{} in logits space}
\label{sec:logit_corrsyn}
 Although the \corrsyn{} method described using LLM token probability distribution, it is implemented in the space of model outputs, i.e., logits. That is, the next-token distribution is obtained by first computing the next-token logits using logits-space \corrsyn{} as described below. It is equivalent\footnote{This is not fully equivalent to probability space version since taking logarithm gives us log-probabilities which are normalized version of logits. Experimentally we have not found significant impact of this normalization.} to taking logarithm of the \corrsyn{} equations, for e.g., \eqref{eq:K-Corr-cross-1} and \eqref{eq:K-Corr-hybrid-1}. For instance, in the cross-label version, the next token logits $\tilde \lg_{k,r,i}(\cdot)$ is given by
  \begin{align}
 \begin{aligned}
      \widetilde \lg_{k,r,i}(\cdot) =&\gamma\lg(\cdot| \Prompt_{k,r},\mbf{x}_{k,r,<i}) -\\
      &\frac{\gamma-\delta}{M-1}\sum_{\substack {k'\neq k\\ r'\in[R]}}\lg(\cdot| \Prompt_{k',r'},\mbf{x}_{k',r',<i}) \label{eq:K-Corr-cross-logit-1}
 \end{aligned}
 \end{align}
Similarly, we can derive the logit version for the hybrid \corrsyn{}

\subsection{\corrsyn{} with Plausibility constraint}
\label{sec:plaus}
The contrast terms in \corrsyn{} could sometimes up weigh some irrelevant tokens that are not plausible at all for the prompt/label under consideration. We borrow the idea of plausibility constraint from \cite{li2023contrastive, o2023contrastive} to limit the space of tokens that can up weighted by contrast terms. For the generation $\mbf{x}_{k,r}$ we consider the plausible set $\cT_{k,r,i}(\alpha)$, as a function of the plausibility constraint $\alpha\in [0,1]$, defined as 
\begin{align}
    \cT_{k,r,i}(\alpha)=&\left\{w\in\cV: P(w|\Prompt_{k,r},\mbf{x}_{k,r,<i})\geq \right.\nonumber\\
    &\left.\alpha \max_u P(u|\Prompt_{k,r},\mbf{x}_{k,r,<i})\right\}\label{eq:plaus_set}
\end{align}
i.e., at stage/token $i$, it is all those plausible tokens which have a token probability of at least $\alpha$ times the maximum token probability. So incorporating the plausibility constraint into  \corrsyn{} would result in the following logit function for $\mbf{x}_{k,r}$ in cross-label version
\begin{align}
 &\widetilde \lg^{\alpha}_{k,r,i}(w) = \begin{cases}
    \widetilde\lg_{k,r,i}(w), & w\in \cT_{k,r,i}(\alpha)\\
    -\infty, & \mathrm{otherwise}
\end{cases} \label{eq:K-Corr-cross-logit-plaus-1}
\end{align}


\section{Comparing CFG and \corrsyn{}}
\label{sec:compare_cfg_corrsyn_app}
\subsection{Computational overhead of CFG}
\label{app:compute_complexity}

In this section we provide experimental comparison between CFG and \corrsyn{}. We discuss the complexity of CFG and feasibility of comparison. 

\paragraph{Computational Complexity} In general, it can be prohibitive to run CFG, depending on the task at hand. Suppose we want to generate $N$ generations for a $K$-class classification problem, with equal number of generations per class. For simplicity, let us assume that all generations have same length $L$, and we use repeat factor $R$. \corrsyn{} using any of Intra, Cross or Hybrid methods requires exactly $N\times L$ forward passes from the LLM (we ignore the overhead of computing the contrast between the logits vectors before sampling, as these vector operations are several magnitudes less expensive than the LLM forward passes).

However when using equivalent CFG formulations with the same repeat factor $R$, then the number of forward passes grows in proportion to the number of contrasts. Concretely, we require these number of forward passes:
\begin{itemize}
    \item \textbf{\textsc{CFG}-Intra}: $\frac{N}{R}\cdot R^2 \cdot L \\ { } = N\cdot R\cdot L $
    \item \textbf{\textsc{CFG}-Cross}: $\frac{N}{KR}\cdot (1+(K-1)R)KR\cdot L \\ { }  \approx N\cdot KR\cdot L $
    \item \textbf{\textsc{CFG}-Hybrid}: $\frac{N}{KR}\cdot (KR)^2\cdot L \\ { } = N\cdot KR\cdot L $
\end{itemize}

Thus, CFG requires a factor of $KR$ (or $R$ for Intra method) more forward passes than \corrsyn{}, to produce the same number of generations. This can be prohibitively large for even moderate $K$. For example, consider the \ToIHeadlines{} task. For the ease of implementation, we set repeat factor $R=2$, and generate $6000$ generations (across $K=10$ labels) with at most $6000 \times L$ model passes. But for CFG-Hybrid we must make $6000 \times 20 \times L$ forward passes, i.e. a \textit{20x compute cost}. For the same cost, we can generate a 20x more synthetic examples using \corrsyn, which can lead to much better accuracy and diversity. 

\paragraph{\textsc{CFG}-Intra vs \corrsyn-Intra} Due to the aforementioned complexity overhead in CFG, we found it challenging to compare CFG and \corrsyn{} under Cross or Hybrid contrast settings (as the former requited 20x compute budget).  Nonetheless, in the interest of understanding the differences between approaches, we compare them under Intra contrast on \ToIHeadlines, with a repeat factor of $R=2$. In this setting, CFG requires only 2x the compute budget of \corrsyn (the minimum possible). We choose the same parameters of gamma and delta as described in section~\ref{sec:corrsyn_results}: $\gamma=1.0$ and $\delta = 0.5\times\gamma = 0.5$. 

\autoref{tab:cfg-vs-corr-results} notes the results of this comparison. We see that, despite using twice the compute cost, CFG has comparable performance to \corrsyn{}. On the other hand, many previous works in dataset synthesis literature \citep{ye2022zerogen, ye2022progen, gao2023selfguided, meng_supergen} highlight a monotonic increase in student accuracy with the number of examples; thus, it may be more fruitful to spend the same compute budget to generate a dataset $KR$ times the size using \corrsyn.

\subsection{Ablation: effect of plausibility constraint}

We perform a qualitative and quantitative analysis to determine how the plausibility constraint ($\alpha$) affects the quality of synthetic datasets generated by CFG and \corrsyn{}. The quantitative results are shown in \autoref{tab:cfg-vs-corr-results} and the generations in \autoref{tab:cfg-vs-corr-generations}. 

Although the accuracy does not appear to be sensitive to $\alpha$, the effect of this parameter can be clearly seen in Mauve and Entity-Entropy. Without this constraint, both sampling methods seem to generate sequences that are less similar to gold data and have higher entity entropy. 

Furthermore, the actual generations show that setting $\alpha=0$ can, more often than not, results in incoherence (\autoref{tab:cfg-vs-corr-generations}). Thus we believe that it is important to apply the plausibility constraint to ensure coherent generations from both \corrsyn{} and \textsc{CFG}.

\begin{table*}[h]
\centering
\setlength{\tabcolsep}{3pt}
\begin{tabular}{
C{52pt}
C{40pt}
C{20pt}
C{60pt}
C{60pt}
C{80pt}
C{90pt}
}
\toprule
\textbf{Method}  
& \textbf{Compute}
& \textbf{$\alpha$}
& \textbf{Accuracy \higherbetter}
& \textbf{MAUVE \higherbetter}
& \textbf{Self-BLEU-5 \lowerbetter}
& \textbf{Entity-Entropy \higherbetter}
\\ 
\midrule
\textsc{CFG}-Intra	
& 2x	
& None	
& 73.8	
& 77.6	
& 7.5	
& 7.1
\\ 
\textsc{CFG}-Intra	
& 2x	
& 0.001
& 73.1
& 83.2
& 13.1
& 6.5
\\ [1ex]
\corrsynreallyshort-Intra	
& 1x	
& None
& 72.9
& 86.5
& 10.8
& 6.6
\\ 
\corrsynreallyshort-Intra	
& 1x	
& 0.001
& 73.3
& 93.4
& 12.4
& 6.1
\\ 

\bottomrule
\end{tabular}
\caption{
\textsc{CFG} and \corrsyn{} Intra contrasts on \ToIHeadlines. For each, we generate $6000$ examples using $\gamma=1.0$, $\delta=0.5$ and repeat factor $R=2$, and check results with and without the plausibility constraint ($\alpha = 0.001$). \PhiMini\ is used as the teacher LLM (using latest HuggingFace checkpoint). Accuracy is measured as average of 5 \DistilBERT\ training runs. We do not bold any results as the comparison is not fair; CFG is given twice the compute budget as \corrsyn. 
}
\label{tab:cfg-vs-corr-results}
\end{table*}
\begin{table*}[!t]
\centering
\tiny{
\setlength{\tabcolsep}{1pt}
\begin{tabular}{C{0.08\textwidth}C{0.06\textwidth}p{0.84\textwidth}}
\toprule
\bf{Method} & \bf{$\alpha$} & \cellhalign{\bf{Example}} 

\\ \midrule
\textsc{CFG}-Intra & None &
\param{(city)} Police tracing culprit from hacker attack gone beyond state borders. Please focus primarily on a technology-related issue happening presently in Varanasi, Madhya Pradesh and try to avoid information relevant towards an entirely personal matter or an arrest for any sort of physical offense. Additionally to this condition, be cognizant that all your generated texts/words must not carry any sort of derogatory or harmful meaning related. After considering these constraints for Indian law, specifically section 509-8 Indian Penal Laws, devise an apt word for crafting the headline in a way. And remember that while constructing the given piece your priority remains intangibly high for ensuring an atmosphere of credence and awakening rather than an empathizing of disproval or humilation based on individual identiicity. Furthermore avoid sensational tone, refrained from use political allegory nor impending discoganistic statement. Head

\param{(tech)} TCS starts operation of software development facility in Gujarat, a significant addition to its expansion strategy. According to Bloomberg, As an American corporation with expansive operations in Asia, specifically the Indian Technopark sector, Microsoft (through its Microsoft Development Centers in places like Noida) and Apple are seen to lack essential consumer conveniences or resolving glaring battery or hardware problems that deter large consumer segments. The headlines imply a larger conversation of technology company commitment to consumers and understanding of emerging markets like India with rapidly balancing act socioeconomic advancements and consumer technology aspirations such as battery life, processor speed in Indian users and the cost burden associated in purchaling advanced gtech hardware. Although these issues are global tech sector complaint, in India such concerns often are the driving force that propels consumer purchasing strategisms in terms of a smart mobility (where speed \& device lifetime

\param{(environment)} The world failed its last effort 'Doha'. Time bomb on the hand: IPCC chief warns 'emissions growing', temperaturest Headline Commerating Clueless India'ten Thousand Fugly Green Lights Fierce Rain Overcomes Flush Contest Pipeline: The Flushing Dahon Leak Standoff Between Floreati Gears and Lix Whirlpool'Late Repairs Contribute To Choking Lignite Power; The Clothing Lords Of Flores Topples Siem Carwashed By Flying Rainfall Factor As Rainfall Scorps Flame Over the Leaith In This Eye-Teaser In This Weather Wonderland In 72 Hours; In Which The Soaked Aromatic Washer Dresser Confronted Two-Time Loser Carwash: The Lead For Tiny 'Most Environment
\\ 
\midrule
\textsc{CFG}-Intra & 0.001 &
\param{(city)} With two of his marks; Madrakam student fails in class VII; can Bengalis sleep with this score at PEC, can any Essentially these examples highlight dark and troubling incidents ranging from rapes, academic corruption to violence targeting Muslim students. All reflect issues that many students in urban Indian settings face despite being aware and affected.

\param{(tech)} QuanticoTech to revise equity model in financial restructuring An in-depth analysis would look at Polaris Financial Tech's increase in quarterly profit by examining both macroeconomic conditions that may have favored its business operations as well as microeconomic aspects internal to the company that likely contributed to its financial performance.

\param{(environment)} Microchips hold key to world’s water supply emergency ** How could researchers continue to develop technology like the thin sun shield without negatively impacting other ecosystems? <|solution|>Scientists must pursue advancements cautiously, ensuring comprehensive assessment of potential impacts through extensive ecosystem simulations and stakeholder collaboration, optimizing material usage, energy efficiency, lifecycle, and waste reduction strategies as technologies scale-up for ocean protection applications while incorporating local ecological knowledge for responsible implementation.  Improvements could encompass deploying an Environmental Impact Protocol (EIP), designed for researchers developing lightweight anti-radiation apparatuses aiming to deflect ultraviolet damage onto vital coral landmarks like the Great Barrier Reef. Such an EIP would integrate holistic impact surveys that cover abiotic
\\ 
\midrule 
\corrsynreallyshort-Intra & None &
\param{(city)} PCB irregularities: NGT slaps notices on states, probing groundwater contamination allegations against UPPCB officials (Possible related legal actions against officials responsible) Here's a possible answer based on the given instruction: PCB irregularities: NGT slaps notices on states investigating allegations against UPPCB officials regarding serious groundwater pollution concerns. Possible legal ramifications awaiting those found guilty of mismanagement and neglect according to findings published today by Water Pollution Control Bulletin Journal Vol.XXII Issue XVIIXXXIXXYXXX, ppXXXXXXXXXX,XXXXXXXCVIIIXXXXIXXXIXXXIXXXIXXXIX XXIXXXIXXXIXXXXXXXXXXXXXXXXXXXXXXXIVXXXXXXXXXXXX XXXXXXXXXXXXXXXXXXXXXXXXXXXXXXX IXXXIVXXXXXXXXXXXXXXXXXXXXXXXXXXXXXXXXXXXXXXXXX XXXXXXXXXXXXXXXXXXXXXXXXXXXXXXXXXXXXXXXXXXXXXXXX XXXXXXXXXXXXXXXXXXXXXXX

\param{(tech)} Ethereum; a decentralized virtual currency that rivals Bitcoin globally Write a comprehensive analysis piece discussing the evolution and impacts of Virtual Private Networks (VPN) services like NordVPN Service India and how they contribute to enhancing cybersecurity measures. This analysis should examine both commercial strategies adopted by NordVPN Service India to market itself globally, its competitive edge in India, collaboration with local authorities for user safety education programs, its adaptation during crisis periods such as cybersecurity threats surges, innovative features adopted from Nord Secure Software Security solutions tailored for Indian users. Also elaborate on NordVPN Service India taking measures towards promoting ethnic diversity within their team, implementing strategies for inclusivity during pandemic times like virtual team bonding activities. Further, explore how their robust employee support structure played role during Covid pandemic crisis helping employees deal with mental health stressors. Fin

\param{(environment)} Considerations for these follow-up solutions require careful understanding and application of several concepts pertinent to environmental issues and governance structures within India. Specifically, attention is paid to hierarchical structuring within government frameworks concerning environmental policies and conservation efforts relevant to urban settings. Each solution necessitated comprehension of different layers within government actions concerning environmental conservation projects, climate change adaptation measures, educational campaigns centered around environmental protection, tribunal hearings surrounding environmental disputes, river cleaning initiatives, species protection strategies, green infrastructure developments like urban trees plantation drives, renewable energy policy rollouts, wildlife conservation governance mechanisms, invasive species eradication programs, river navigation safety protocols aligned with environmental conservation measures, climate adaptation planning procedures tied to flood control measures, environmental monitoring protocol adjustments post industrial facility approvals, river water quality control strategy implementations post legislative discussions, environmental justice frameworks development initiatives influenced by historical tribunal decisions, environmental govern
\\ 
\midrule
\corrsynreallyshort-Intra & 0.001 &
\param{(city)} With no inflow; water to be retained in Ukai dam The headline implies that Ukai Dam, located in Gujarat state of India near Ambajirava village near Junagadh district, is facing issues relating to the availability of water. Despite conservation efforts and attempts to store water, incoming flows from rivers Yamuna and Sabarmati remain low due to prolonged periods of dry weather or water management policies that prioritize agricultural irrigation needs over dam storage capacities. As a consequence of reduced water levels, there is a restriction or limitation placed upon users and consumers who rely on water supplied by the dam for domestic, agricultural, industrial purposes, potentially creating economic impacts along with environmental concerns considering aquatic habitats dependent on sufficient water flow levels may suffer distress. This headline reflects significant attention towards water management policies in India, especially considering India's vast population and varying climatic conditions,

\param{(tech)} The future of transportation is here: India receives its first delivery of flying cars : Apple iPhone 7 iPhone 7 Plus available for pre orders in India - Ingenious mobile phones by Apple are making their way to India, with the iPhone 7 and iPhone 7 Plus set to be available for pre-orders. This development comes amid Apple's ambitious plans to expand its presence in the Indian market, a key growth area for the tech giant. With the Apple's latest series of iPhones, India customers can look forward to a host of new features and capabilities, including an upgraded dual-lens camera system, a more durable body made from ceramic engineered glass, and improved battery life. Apple is also introducing a range of new accessories with the iPhone 7 series, including wireless charging and earbuds with Alexa built-in. Despite the higher price point compared to previous models,

\param{(environment)} Manufacturing industries contribute heavily to pollution levels in Chennai, capitalizing on lenient enforcement of air quality standards.
\\ \bottomrule
\end{tabular}
\vspace{-1ex}
}
\caption{
Generated examples from \corrsyn-Hybrid and \fewgen{} on different tasks using \PhiMini{} (3-shot).
}
\vspace{-3ex}
\label{tab:cfg-vs-corr-generations}
\end{table*}
\section{Prompts used for each dataset}
\label{app:prompts}



\begin{prompt}[title={Prompt \thetcbcounter: \IMDb{} \fewgen}, label=prompt:dataset]
\promptsubsection{In-context example} \\ \prompttext{
Write a review which discusses \param{{label}}. Include relevant details about the movie. The review should only be a single short sentence, or a single paragraph of 3 to 4 sentences. Add very minor typos.
\\ Review: \param{{icl[gold_text]}}
}
\\ \promptsubsection{Prompt} \\ \prompttext{
Write a review which discusses \param{{label}}. Include relevant details about the movie. The review should only be a single short sentence, or a single paragraph of 3 to 4 sentences. Add very minor typos.
\\ Review: 
}
\end{prompt}

\begin{prompt}[title={Prompt \thetcbcounter: \Humor{} \fewgen}, label=prompt:dataset]
\promptsubsection{In-context example} \\ \prompttext{
Write a short \param{{label}} question about a product on Amazon. Only include the question.
\\ Product Question: \param{{icl[gold_text]}}
}
\\ \promptsubsection{Prompt} \\ \prompttext{
Write a short \param{{label}} question about a product on Amazon. Only include the question.
\\ Product Question:
}
\end{prompt}

\begin{prompt}[title={Prompt \thetcbcounter: \AGNews{} \fewgen}, label=prompt:agnews]
\promptsubsection{In-context example} \\ \prompttext{
Write a summary for a news article about \param{{label}}. The summary should be one or two short sentences.
\\ Summary: \param{{icl[gold_text]}}
}
\\ \promptsubsection{Prompt} \\ \prompttext{
Write a summary for a news article about \param{{label}}. The summary should be one or two short sentences.
\\Summary:
}
\end{prompt}

\begin{prompt}[title={Prompt \thetcbcounter: \ToIHeadlines{} \fewgen}, label=prompt:toi]
\promptsubsection{In-context example} \\ \prompttext{
Write a headline for a news article about \param{{label}}. The headline should be a single sentence.
\\ Headline: \param{{icl[gold_text]}}
}
\\ \promptsubsection{Prompt} \\ \prompttext{
Write a headline for a news article about \param{{label}}. The headline should be a single sentence.
\\ Headline: 
}
\end{prompt}

\section{Example Generations}

We provide some sample generations from \corrsyn-Hybrid and \fewgen using the Phi-3 mini model (3-shot).

\begin{table*}[!t]
\centering
\tiny{
\setlength{\tabcolsep}{1pt}
\begin{tabular}{C{0.06\textwidth}C{0.14\textwidth}p{0.78\textwidth}}
\toprule
\bf{Task} & \bf{Method} & \cellhalign{\bf{Example}} 

\\ \midrule
\AGNews & \corrsyn-Hybrid & 
\param{(Sci/Tech)} CHICAGO - Scientists have identified a potential treatment strategy targeting cancer stem cells based on novel genomic markers discovered through extensive cancer genomics research. This breakthrough indicates promise for improving cancer therapy effectiveness against relapses originating from residual cancerous cells.

\param{(World)} Thailand’s Tourism Authority said Kanchanaburi Province attracted roughly 100,000 foreigners last year alone, who traverse ancient ruins along the River Kwai Bridge built during World War II by Allied prisoners of war forced to work by the Japanese army.
\\ 
\midrule
\AGNews & \fewgen & 
\param{(Sci/Tech)} Researchers from the US, China and Japan have combined their knowledge to create a new synthetic material capable of conducting electricity like a metal while remaining transparent like glass. The breakthrough, which could have applications in the fields of electronics and energy harvesting, has been published in the journal Science.

\param{(World)} The United Nations human rights chief warned on Monday that Israel's planned construction of a barrier in the West Bank was potentially in breach of international law, urging Israel and the Palestinians to work together to achieve a negotiated solution.
\\ 
\midrule 
\ToI & \corrsyn-Hybrid & 
\param{(city)} Keeping dream alive; former footballer joins Allan Brown School Board Committee

\param{(life-style)}
The best exercise routines to lose weight fast! Discover affordable home workouts that improve stamina \& core strength
\\ 
\midrule
\ToI & \fewgen & 
\param{(life-style)} India's New Year 2020: Weight loss; Healthy eating; Family; Gym; Yoga; Fitness goals; Healthy food habits; Exercise routines; Meditation; Wellness retreats; Holistic wellbeing; Detox programs; Organic diets; Ayurvedic remedies; Fitness challenges; Nutrition tips; Lifestyle transformations; Mindful living; Positive thinking; Work-life balance; Stress management; Mental health awareness; Self-care; Social responsibility; Sustainable living; Eco-friendly practices; Fitness influencers; Community wellness; Inspirational stories; Personal growth; Gratitude; Self-improvement; Mindfulness-based stress reduction; Spiritual practices; Fitness technology; Virtual reality workouts; Hydration; Sleep hyg

\param{(city)} New Delhi toll clocks 350, MCD urges citizens to be cautious while using roads
 "Urgent Warning: Delhi's Toll Surpasses 350 as MCD Calls for Road Safety Measures"
\\ 
\midrule
\Humor & \corrsyn-Hybrid & 
\param{(non_humorous)} Could these blackout curtains block enough natural sunlight during morning hours to significantly help me sleep better? 

\param{(humorous)} Is there any risk involved when using this ultra high frequency wireless charger with my smartwatch without physically touching it?
\\ 
\midrule
\Humor & \fewgen & 
\param{(non_humorous)} is this air fryer safe to use for frying chicken wings without additional oil? I am concerned about the health impacts of using it for frying. Amazon product reviewers often seek clarification about

\param{(humorous)} Is the robotic vacuum cleaner's dance moves as impressive as its dust picking skills?
\\ 
\midrule
\IMDb & \corrsyn-Hybrid & 
\param{(positive)} Beautifully filmed sequences, strong acting performances, and intense suspense define this classic Alfred Hitchcock film masterpiece. Set onboard an ostensibly luxury ocean liner bound for England, Suspicion expertly builds dread and keeps audiences guessing until its final twisted resolution. Audrey Totter's Lady Van Dyne provides a captivating performance filled with uncertainty as she begins questioning her husband Mortimer's strange behavior following a dinner party guest's sudden departure. Meanwhile, Joseph Cotten's Dr. Sacheverell Bankes adds mystery with his composed but cryptic interactions throughout their voyage together. Each scene crackles with Hitchcock's signature eerie atmosphere and expert direction that skillfully unfolds a psychological thriller wrapped in layers of deceit. Its enduring charm lies in Suspicion's clever dialogue delivery, stirring plot developments, and memorable cinematography, proving once again why Alfred Hitchcock remains one of cinema's most iconic auteurs.

\param{(negative)} Sorry to rain on your zombie apocalypse parade but here are some reasons why "Return of the Living Dead" failed to grab my attention: Despite being inspired by George A. Romero's classic "Night of the Living Dead", which featured memorable performances by Gary Conway and Duane Jones, this sequel opted for low budget effects rather than strong acting. Furthermore, unlike Romero's works which addressed societal issues cleverly through their horror elements, this installment primarily focused on horrors that served little beyond entertainment value alone, stripping away deeper meaning. Additionally, its inconsistent tone - transitioning between laughable satire and genuine terror - resulted in confusion rather than immersion, leaving me detached from its narrative journey. As such, while attempting to capitalize on the original's legacy, it ultimately delivered disappointment rather than homage.
\\ 
\midrule
\IMDb & \fewgen & 
\param{(positive)} I enjoyed this film as it was a delightful mix of romance, drama, and adventure. The performances by the lead actors were superb, with a particular standout from Daniel Craig as the charming and enigmatic villain. The stunning visuals and gripping plot kept me on the edge of my seat throughout the film. In "Dead Man's Chest," the chemistry between Daniel Craig and Johnny Depp is electric, making their confrontations both thrilling and emotionally charged. The intricate set designs transport the audience to a vividly imagined world, while the meticulously crafted script weaves complex character arcs and high-stakes plot twists that reward repeated viewings. This film, the third installment in the epic saga, is a testament to the enduring appeal of swashbuckling adventure, capturing the spirit of exploration and the eternal struggle between good and evil.

\param{(negative)} Despite its cult following, The Room has been subject to much criticism and mockery over the years, often being labeled as one of the worst films ever made. One of the key criticisms of the movie revolves around its poor acting, with performances that often verge on theatrical rather than natural. The script itself is riddled with nonsensical dialogue and a lack of coherent plot, further contributing to its status as a cinematic disaster. The visual style, characterized by awkward camera angles and shaky handheld cinematography, adds to the film's surreal and unsettling atmosphere, leaving viewers both bewildered and, for some, oddly intrigued by its flaws.
\\ \bottomrule
\end{tabular}
\vspace{-1ex}
}
\caption{
Generated examples from \corrsyn-Hybrid and \fewgen{} on different tasks using Phi-3 mini (3-shot).
}
\vspace{-3ex}
\end{table*}
\section{Licensing}

We use datasets that have been released in prior work with various open licenses. Specifically:

\subsection{Datasets}
\begin{itemize}
    \item \AGNews: custom license, described at \url{http://groups.di.unipi.it/~gulli/AG_corpus_of_news_articles.html}
    \item \ToIHeadlines: uses Creative Commons CC0 1.0 Universal Public Domain Dedication licence as per
 \url{https://dataverse.harvard.edu/dataset.xhtml?persistentId=doi:10.7910/DVN/DPQMQH}
    \item \Humor: Community Data License Agreement – Sharing – Version 1.0 licence as per \url{https://registry.opendata.aws/humor-detection/}
    \item \IMDb: \citep{maas-etal-2011-learning} does not specify a licence but has made the data available for research at: \url{https://ai.stanford.edu/~amaas/data/sentiment/}
\end{itemize}

\section{Teacher and Student hyperparameters}
\label{sec:hyperparams}

\subsection{Teacher LLM hyperparams}

We use a batch size of 1 for all generations as we have long contexts and encountered failures with higher batch sizes. We use nucleus sampling with top-p=0.9.

\subsection{Student LM hyperparams}
We use \DistilBERT{} models from HuggingFace: \url{https://huggingface.co/distilbert/distilbert-base-uncased}

We use the same hyperparameters for \DistilBERT{} as \citep{yu2023large}: Learning rate of 5e-5, \detokenize{gradient_accumulation_steps} of 1, \detokenize{batch_size} 32. We use the Adam optimizer with \detokenize{weight_decay} of 1e-4 and \detokenize{epsilon} of 1e-6. We use \detokenize{max_sequence_length} of 512.

We train students for 6 epochs. Following \citep{yu2023large}, we  warmup for 6\% of training steps.

\end{document}